%% file: neurips_2025.tex
\documentclass{article}

    \PassOptionsToPackage{numbers, compress}{natbib}
\usepackage[final]{neurips_2025}

\usepackage[utf8]{inputenc} % allow utf-8 input
\usepackage[T1]{fontenc}    % use 8-bit T1 fonts
\usepackage{hyperref}       % hyperlinks
\usepackage{url}            % simple URL typesetting
\usepackage{booktabs}       % professional-quality tables
\usepackage{amsfonts}       % blackboard math symbols
\usepackage{nicefrac}       % compact symbols for 1/2, etc.
\usepackage{microtype}      % microtypography
\usepackage{xcolor}         % colors
\usepackage{xspace}
\usepackage{graphicx}
\usepackage{subcaption}
\usepackage{wrapfig}
\usepackage{amsmath}
\usepackage{multirow}
\setcitestyle{numbers}
\usepackage[table,xcdraw]{xcolor} % table use
\usepackage{makecell}

\usepackage{bm}
\usepackage{algorithm}
\usepackage{listings}
\usepackage{tabularx}
\usepackage{enumitem}
\definecolor{codeblue}{rgb}{0.25, 0.5, 0.5}
\definecolor{codekw}{rgb}{0.35, 0.35, 0.75}
\lstdefinestyle{Pytorch}{
    language         = Python,
    backgroundcolor  = \color{white},
    basicstyle = \fontsize{8.0pt}{9pt}\selectfont\ttfamily\bfseries,
    columns          = fullflexible,
    breaklines       = true,
    captionpos       = b,
    commentstyle     = \fontsize{4pt}{4pt}\color{codeblue},
    keywordstyle     = \fontsize{4pt}{4pt}\color{codekw},
    morekeywords     = {with,scatter_,norm,sort},
}
\newcommand{\paratitle}[1]{\vspace{1.5ex}\noindent\textbf{#1}}
\newcommand{\ie}{\emph{i.e.,}\xspace}

\newcommand{\eg}{\emph{e.g.,}\xspace}
\newcommand{\etal}{\emph{et al.}\xspace}

\newcommand{\ignore}[1]{}
\usepackage{tcolorbox}
\definecolor{darkorange}{RGB}{255, 140, 0}
\definecolor{darkblue}{RGB}{84, 112, 198}
\definecolor{lightgreen}{RGB}{145, 204, 117}
\definecolor{lightyellow}{RGB}{250, 200, 88}
\definecolor{lightred}{RGB}{238, 102, 102}
\definecolor{lightblue}{RGB}{115, 192, 222}
\newtcolorbox{promptbox}[2][Prompt]{
colback=black!5!white,
arc=5pt, 
boxrule=0.5pt,
fonttitle=\bfseries,
title=#1, 
before upper={\normalsize}, fontupper=\fontfamily{ptm}\selectfont,
colframe=#2, 
}
\title{
\fontsize{15pt}{17pt}\selectfont
How Efficient Are Diffusion Language Models?\\
A Critical Examination of Efficiency Evaluation Practices
}

\author{Han Peng$^{1}$,~Peiyu Liu$^{2}$,~Zican Dong$^{1}$,~Daixuan Cheng$^{1}$,~Junyi Li$^{4}$,~Yiru Tang$^{1}$,\\~\textbf{Shuo Wang}$^{3}$,~\textbf{Wayne Xin Zhao}$^{1}$\thanks{Corresponding author.}\\
        $^1$ Gaoling School of Artificial Intelligence, Renmin University of China\\
        $^2$ University of International Business and Economics ~~$^3$ Tsinghua University \\
        $^4$ Department of Data Science, City University of Hong Kong \\
        \texttt{panospeng@ruc.edu.cn, batmanfly@gmail.com}\\
}

\begin{document}

\maketitle

\begin{abstract}
Diffusion language models (DLMs) have emerged as a promising alternative to the long-dominant autoregressive (AR) paradigm, offering a parallelable decoding process that could yield greater efficiency. Yet, in practice, current open-source DLMs often underperform their AR counterparts in speed, limiting their real-world utility. This work presents a systematic study of DLM efficiency, identifying key issues in prior evaluation methods. Through empirical benchmarking and a theoretical analysis, we demonstrate that AR models generally achieve higher throughput, while DLMs consistently lag. We also investigate acceleration strategies, finding that techniques like dual cache and parallel decoding mainly offer  gains at small batch sizes, with their benefits diminishing upon scaling. Our findings underscore the necessity of robust evaluation methods and improved acceleration strategies to advance research on DLMs.
%clarify the practical efficiency of DLMs and underscore the open challenges in making diffusion-based language modeling truly competitive. 
\end{abstract}

\input{sections/introduction}
\input{sections/background}
\input{sections/evaluation_issues}
\input{sections/desigh_setup}
\input{sections/evaluation_analysis}

\input{sections/future_directions}
\input{sections/conclusion}

%\subsubsection{Case Study}

\bibliographystyle{unsrt}
\bibliography{ref.bib}

%%%%%%%%%%%%%%%%%%%%%%%%%%%%%%%%%%%%%%%%%%%%%%%%%%%%%%%%%%%%
\input{sections/appendix}

% \newpage
% \input{checklist}
% \appendix
% \input{appendix}

%%%%%%%%%%%%%%%%%%%%%%%%%%%%%%%%%%%%%%%%%%%%%%%%%%%%%%%%%%%%

\end{document}

%% file: sections/introduction.tex
\section{Introduction}
\label{sec:intro}
Diffusion language models (DLMs) have recently emerged as a competitive alternative to the long-dominant autoregressive (AR) approach~\cite{llmsurvey-arxiv-2023-wayne} for natural language generation. Despite the relatively low profile of non-autoregressive research, its persistent advancement has now yielded performant models like LLaDA~\cite{dllm-arxiv-2025-shen} that are competitive with their AR counterparts. This represents a significant departure from the sequential generation paradigm that has defined the field for years.

In contrast to AR models, which generate text one token at a time, DLMs produce multiple tokens in parallel via an iterative denoising process. This approach theoretically offers a path to overcome the efficiency bottleneck of sequential generation in AR models. In practice, however, a ``paradox'' has become increasingly evident: current open-source DLMs often demonstrate slower inference speeds than their AR counterparts of similar scale. For example, LLaMA3-Instruct-8B \cite{llama3-arxiv-2024-abhimanyu} exhibits an inference throughput 13.7$\times$ greater than that of LLaDA-Instruct-8B on evaluation benchmarks \cite{4-arxiv-2025-xu}. This ``theoretically fast but practically slow'' predicament greatly limits the deployment of DLMs in real-world applications, as their potential speed advantages are negated by implementation inefficiencies.

To improve the inference efficiency of DLMs, recent research has explored various acceleration strategies, which can broadly fall into two categories: (1) reducing per-step computational overhead (\eg using lighter diffusion steps or more efficient sampling methods) and (2) reducing the number of diffusion sampling steps required (\eg through faster convergence or distillation). 

Despite this progress, our extensive review of the literature identifies a fundamental problem: a widespread lack of rigorous evaluation for efficiency improvements in DLMs. Specifically, we find three major issues in existing research on acceleration and evaluation of DLMs: 

\vspace{-0.2cm}
\begin{itemize}[leftmargin=0.7cm]
    \item \textbf{Evaluation Scope:} Prior evaluations often compare DLMs and AR models in limited conditions, such as single instance inference or fixed generation length. These constrained conditions fail to capture performance\footnote{In this paper, \textit{performance} refers to system-level efficiency metrics rather than task-level or downstream accuracy.} across diverse settings and poorly represent real-world use.
    
    \item \textbf{Infrastructure Implementation:} Some impressive studies rely on hybrid or proprietary infrastructure implementations, mixing algorithmic innovations with kernel- or system-level optimizations, which blurs the boundary between algorithmic and engineering gains.
    
    \item \textbf{Efficiency Metrics:} Reported metrics are inconsistent across studies, ranging from latency per token to throughput on specific hardware, making it difficult to compare results or account for true computational cost.
\end{itemize}
\vspace{-0.2cm}

To better promote the development of DLMs, this work aims to systematically evaluate their efficiency and investigate their performance relative to AR models across a wider range of conditions. Our investigation is guided by three key research questions:  

\vspace{-0.2cm}
\begin{itemize}[leftmargin=0.7cm]
    \item \textbf{RQ1: How do DLMs compare with AR Models in efficiency?}
    
    \item \textbf{RQ2: How can we theoretically analyze the variations in inference throughput across architectures?}
    
    \item \textbf{RQ3: How do acceleration strategies benefit DLMs in varied conditions?}
\end{itemize}
\vspace{-0.2cm}

Guided by these questions, we undertake a further investigation, integrating empirical analysis with theoretical insights into DLM efficiency and acceleration. The principal findings and contributions of this work are as follows: 

\vspace{-0.2cm}
\begin{itemize}[leftmargin=0.7cm]
    \item \textbf{Systematic efficiency evaluation:} We conduct a comprehensive comparison of three model types—DLM,  AR and block diffusion—under varying sequence lengths and batch sizes. Our findings show that AR models consistently achieve the highest throughput, followed by block diffusion, with DLMs being the slowest across most evaluated settings, including different prompt lengths, generation lengths and batch sizes.
    
    % \item \textcolor{blue}{\textbf{Throughput modeling and theoretical analysis:} We perform a theoretical analysis that decomposes inference throughput into hardware-side compute capacity (FLOPs/s) and model-side compute efficiency (FLOPs/token). Based on the roofline model, we analytically characterize the throughput behavior of autoregressive, diffusion, and block diffusion models under varying sequence lengths and batch sizes.}
    % 执行理论分析，目的是分析什么，我们要做什么事情
    % \item \textcolor{blue}{\textbf{Throughput modeling and theoretical analysis:} We perform a theoretical analysis based on established efficiency modeling frameworks, viewing throughput as a function of hardware efficiency~(FLOP/s) and model-side workload~(FLOPs/token). To this end, we analytically characterize the throughput behavior of autoregressive, diffusion, and block diffusion models under varying sequence lengths and batch sizes.}
    \item {\textbf{Throughput modeling and theoretical analysis:} We conduct a theoretical analysis aiming to provide a fine-grained understanding of inference throughput. Specifically, we model throughput as a function of \textbf{hardware-side performance}~(FLOPs/s) and \textbf{model-side efficiency}~(FLOPs/token), and analytically characterize how autoregressive, diffusion, and block diffusion models behave under varying sequence lengths and batch sizes.}
    
    \item \textbf{Empirical insights into acceleration strategies:} We analyze two major types of acceleration methods for DLMs—reducing per-step cost (\eg dual cache) and reducing step count (\eg parallel decoding). We find that these acceleration strategies yield significant gains at a batch size of 1, sometimes outperforming AR models, but their advantage diminishes as batch size grows, eventually falling behind AR.
\end{itemize}
\vspace{-0.2cm}

The remainder of this technical report is organized as follows: Section~\ref{sec:background} reviews relevant background and related work. Section~\ref{sec:evaluation_issues} discusses key issues in current efficiency evaluation practices. Section~\ref{sec:desigh_setup} introduces our three core research questions and details the experimental setup. Section~\ref{sec:evaluation_analysis} presents our main empirical results and theoretical analysis. Finally, Section~\ref{sec:future_directions} outlines future directions and open challenges for advancing efficient diffusion-based language modeling.

%% file: sections/background.tex
\section{Background and Related Work}
\label{sec:background}

In this section, we review the background and related work about different paradigms of language models.  

\subsection{Inference Paradigms of AR, Diffusion, and Block Diffusion Language Models}
\label{sec:inference_paradigms}

\subsubsection{Autoregressive Language Models}
Autoregressive language models represent the dominant paradigm in current large language models, such as GPT, LLaMA, and Qwen series. They are trained to predict the next token by attending only to previous ones, which yields high generation quality but enforces strictly sequential decoding. To speed up this process, the \textbf{KV Cache} stores key and value tensors from previous tokens, eliminating redundant computation and enabling faster inference. However, this comes at the cost of significantly increased memory usage and bandwidth pressure.

\subsubsection{Diffusion Language Models}
Diffusion language models generate text through an iterative denoising process: starting from random noise, they progressively refine the sequence into a coherent sample from the data distribution. Unlike AR models, DLMs update all token positions in parallel during each refinement step, allowing for simultaneous generation. Architecturally, they typically use \textbf{bidirectional (non-causal) attention}, allowing every token to attend to the full context of the entire sequence. However, vanilla bidirectional attention is not compatible with KV caching, leading to higher latency as sequence length grows. A notable example of DLMs is LLaDA, a masked diffusion model trained from scratch. It employs a forward data masking process and a reverse process parameterized by Transformer to predict masked tokens, demonstrating strong scalability and competitive performance on various benchmarks.

\subsubsection{Block Diffusion Language Models}
Block diffusion models combines autoregressive dependencies across blocks with parallel diffusion refinement within each block, maintaining long-range contextual coherence while enabling efficient parallel generation. This architecture also naturally supports KV caching, enhancing inference throughput. It introduces a complementary attention mask that enables bidirectional attention within blocks while preserving autoregressive dependencies across them. During inference, this design allows the model to perform autoregressive generation across blocks and parallel diffusion refinement within each block, leading to a significant reduction in training cost and improved inference efficiency. For example, BD3-LM~\cite{5-iclr-2025-marianne}, trained from scratch, demonstrates that block-level diffusion can substantially improve efficiency without sacrificing generation quality. Building on this direction, D2F and Fast-dLLM v2~\cite{6-arxiv-2025-chengyue} show that block diffusion-style generation can also be achieved by training from existing models—D2F from diffusion backbones and Fast-dLLM v2 from autoregressive ones.

\subsection{Strategies for Efficient DLM Inference}
\label{sec:efficient_strategies}
Existing acceleration methods for DLMs can be broadly categorized into two approaches: reducing per-step computational overhead and reducing the number of diffusion sampling steps.

\subsubsection{Reducing Per-step Computational Overhead}

Representative approaches for reducing per-step cost can be classified into two lines: 

\paratitle{Sequence-level KV Cache.} One line of works store and reuse key/value states across decoding steps: The conventional KV cache is designed for strictly autoregressive decoding and does not directly apply to DLMs. To address this, recent work shows that redesigning the decoding schedule allows DLMs to recover much of the efficiency benefit of KV caching. For example, Fast-dLLM~\cite{7-arxiv-2025-chengyue} introduces a DualCache (prefix + suffix) to improve reuse while bounding quality loss. DPad~\cite{8-arxiv-2025-xinhua} proposes a suffix-window and distance-decay dropout to restrict attention and thereby limit suffix token caching and computation. And Sparse‑dLLM~\cite{9-arxiv-2025-yuerong} develops a training-free dynamic cache eviction framework that retains only salient token states in cache and evicts less relevant prefix/suffix entries to improve throughput.

\paratitle{Step-level Feature Reuse.} Another line of works reuse stable intermediate representations inside each denoising iteration. dKV-Cache~\cite{10-arxiv-2025-xinyin} introduces a delayed caching scheme, where a token’s key and value states are cached one denoising step after it is decoded. dLLM-Cache~\cite{21-arxiv-2025-zhiyuan} uses feature similarity to identify stable tokens and reuse their cached features., and FreeCache~\cite{11-arxiv-2025-zhanqiu} reuses stable KV states of early clean tokens while updating only actively changing ones. These methods are complementary to sequence-level caching.

\subsubsection{Reducing Diffusion Sampling Steps}

Another major direction for improving DLM efficiency is reducing the number of denoising steps required during generation. Existing methods typically achieve this through parallel decoding or progressive distillation:

\paratitle{Parallel Decoding.} The main challenge is that token predictions interfere with each other because they are dependent. Fast-dLLM counters this by unmasking tokens whose predicted probabilities exceed a confidence threshold. Tokens below the threshold remain masked until later rounds.

\paratitle{Step Reduction via Distillation.} A key direction for accelerating DLMs is reducing the number of denoising steps through distillation. Early works on diffusion models, such as Progressive Distillation~\cite{12-iclr-2022-tim}, adopt teacher–student schemes that iteratively halve the required sampling steps. Di4C~\cite{13-arxiv-2024-satoshi} extends this approach to discrete diffusion by explicitly distilling inter-token correlations. Recently, DLM-One~\cite{14-arxiv-2025-tianqi} apply these distillation ideas to continuous DLMs, using score-based distillation to train a one-step generator that produces the full sequence in a single forward pass, achieving substantial speedups while maintaining near-teacher quality.

%% file: sections/evaluation_issues.tex
\section{Efficiency Evaluation Issues}
\label{sec:evaluation_issues}

While acceleration techniques for DLMs are advancing rapidly, the field currently lacks standardized evaluation protocols. The wide divergence in experimental configurations across studies often renders their efficiency claims incomparable—or even an inaccurate reflection of genuine performance gains. By reviewing existing evaluation methodologies, we identify several issues that may affect the comprehensiveness of efficiency evaluation. Our objective is to highlight these challenges in efficiency evaluation, thereby fostering more rigorous and comparable research practices within the field. 

% \subsection{Evaluation Scope}
\paratitle{Evaluation Scope.}
Current evaluations of these techniques primarily focus on simplified and constrained settings—most notably using a batch size of one and fixed output lengths. Such limited scope fails to capture the full range of real-world deployment scenarios: evaluations cannot fully reflect how these methods perform across varying batch sizes, output length distributions, or diverse generation tasks. The lack of comprehensive benchmarking across different operational conditions limits our understanding of when and where each acceleration technique provides genuine advantages.

% \subsection{Infrastructure Implementation}
\paratitle{Infrastructure Implementation.}
Another challenge lies in the heterogeneous infrastructure configurations adopted across studies. Fair efficiency comparisons require controlled inference environments, yet many works integrate kernel- or system-level optimizations into their core modeling contributions. Some representative DLMs that report impressive decoding speeds remain closed-source, making it difficult to verify whether the gains arise from architectural innovations or implementation optimizations. Such inconsistencies hinder reproducibility and obscure the true sources of efficiency improvements across the field.

% \subsection{Efficiency Metrics}
\paratitle{Efficiency Metrics.}
Efficiency reporting are inconsistent across studies, making it difficult to compare models fairly. Some studies report latency per token~\cite{11-arxiv-2025-zhanqiu}, while others focus on throughput (tokens per second) on specific hardware setups~\cite{9-arxiv-2025-yuerong, 10-arxiv-2025-xinyin}. These metrics, while useful, often fail to capture the complete picture. For example, a model might show high throughput, but if each step involves heavy computation, the overall cost could still be high. This is especially relevant for DLMs, which often require more operations per token than AR models. To support clearer and more practical comparisons, it would be helpful to include standardized metrics that consider both computational cost and decoding performance. Examples might include FLOPs per token or throughput under fixed resource constraints. These indicators can offer a more balanced view of model efficiency across different architectures.

%% file: sections/desigh_setup.tex
\section{Experimental Design and Setup}
\label{sec:desigh_setup}

\subsection{Overview and Research Questions}
Motivated by the limitations of existing works discussed in Section 3, this paper presents an empirical study of inference efficiency for current architectures. We conduct a comparative analysis across three foundational architectures AR, DLM and block diffusion—along with their respective acceleration methods. Specifically, we organize this study around three key research questions. For each question, we perform targeted experiments and analyses to help readers develop a clearer understanding of the efficiency differences. The three research questions under investigation are listed as follows:

\paratitle{RQ1: How do DLMs compare with AR Models in efficiency?} 
We conduct a controlled comparison of these architectures, evaluating their throughput under varying conditions. The analysis specifically investigates the impact of critical factors such as prompt length, generation length, and batch size.

\paratitle{RQ2: How can we theoretically explain variations in inference efficiency?} 
% \textcolor{blue}{We aim to provide a theoretical perspective by decomposing throughput into effective hardware performance (FLOPs per second) and model-specific workload (FLOPs per generated token). This will provide a clear explanation for the empirical results in RQ1, identify key computational bottlenecks, and suggests a pathway for developing future acceleration methods.}
{We aim to provide a fine-grained theoretical analysis that interprets throughput in terms of hardware-side performance~(FLOPs/s) and model-side efficiency~(FLOPs/token). This analysis helps explain the empirical results in RQ1, reveal key computational bottlenecks, and suggest potential directions for future acceleration.}
% This provides an explanation for the empirical results in RQ1, identify key computational bottlenecks, and suggests a pathway for developing future acceleration methods.}

\paratitle{RQ3: How do DLMs benefit from acceleration strategies under varied conditions?} 
In RQ3, we investigate existing acceleration strategies by classifying them into two main categories: reducing the computational cost per denoising step and decreasing the total number of steps. We evaluate their performance across varied inference scenarios to reveal conditions under which these strategies yield meaningful efficiency gains and guide the design of more efficient DLMs.

\subsection{Experimental Setup}
To ensure a fair and consistent comparison of inference efficiency across different models and methods, we standardized our experimental setup as follows:

\subsubsection{Inference Framework and Hardware}
All evaluations are conducted using the \texttt{Hugging Face Transformers} library. 
To maintain consistency in core computations, the attention mechanism for all models 
leverages PyTorch's official \texttt{torch.nn.functional.scaled\_dot\_product\_attention}. 
All experiments are conducted on a single NVIDIA A800 GPU (80~GB) using FP16 precision. 
The experiments are run within a unified Conda virtual environment to ensure that all 
core dependencies, including PyTorch and CUDA, are identical across all tests.

\subsubsection{Benchmark and Scenarios}
We use the GSM8K dataset for all performance benchmarks. 
The prompt length is controlled by adjusting the number of examples 
(\ie 0-shot and 5-shot) in the prompt. 
For generation length, DLMs allow direct parameter control. 
For AR and block diffusion models, we enforce generation to continue 
until the target length is met (\ie ranging from 64 to 2048), 
even after an ``<eos>'' token is produced, to ensure a fair 
comparison of throughput.

\subsubsection{Models and Acceleration Methods}
\label{sec-models}

We evaluate three representative models and their default decoding strategies:

\vspace{-0.2cm}
\begin{itemize}[leftmargin=0.7cm]
    \item \textbf{LLaDA-8B-Instruct (DLM)\footnote{{https://huggingface.co/GSAI-ML/LLaDA-8B-Instruct}}:} it achieves optimal performance when the number of sampling steps matches the output length, making each decoding step effectively generate about one token, without KV-cache support.
    
    \item \textbf{LLaMA-3.1-8B-Instruct (AR)\footnote{{https://huggingface.co/meta-llama/Llama-3.1-8B-Instruct}}:} it follows a left-to-right autoregressive decoding process, generating one token per decoding step with KV-cache.
    
    \item \textbf{Fast-dLLM v2-7B~(block diffusion)\footnote{{https://huggingface.co/Efficient-Large-Model/Fast\_dLLM\_v2\_7B}}:} it generates text block by block—each block decoded sequentially and predicted in parallel internally—but, similar to LLaDA, each decoding step effectively produces about one token, while supporting KV-cache.
\end{itemize}
\vspace{-0.2cm}

Specifically for the LLaDA model, we evaluate two acceleration strategies introduced in the Fast-dLLM series: the dual cache strategy (to reduce per-step computation) and the confidence-aware parallel decoding strategy (to reduce the number of steps). For the block diffusion, which inherently supports KV cache utilization due to its block-wise autoregressive design, we analyze the confidence-aware parallel decoding strategy. All hyperparameters for these acceleration methods are configured according to their original papers' settings for GSM8K, and we further verify that the accuracy remains aligned with the original models.

\subsubsection{Evaluation Metrics}
We mainly consider two commonly used evaluation metrics:

\paratitle{Throughput.}
To quantify time efficiency, we report the end-to-end decoding throughput in tokens per second. This metric is computed by dividing the number of generated tokens by the total wall-clock time, measured from the start of the generation process to the completion of the final token.

{\paratitle{Arithmetic Intensity.}}
In model inference, the overall latency is determined by both computational workload and memory access. The computational workload is typically measured in floating-point operations (FLOPs), while memory activity is quantified in bytes of read and write operations (MOPs). Furthermore, the arithmetic intensity (Luebke \etal~\cite{gpgpu-siggraph-2004-david}) is defined as the ratio of floating-point operations to memory operations, measuring the balance between computation and data movement: 

\begin{equation}
    \text{Arithmetic Intensity} = \frac{\text{Volume\ of\ Computation (FLOPs)}}{\text{Volume\ of\ Memory\ Access (MOPs)}} \quad (\text{FLOPs/Byte})
\end{equation}

Arithmetic intensity characterizes how many floating-point operations are executed for each byte of data moved. 
A low arithmetic intensity indicates that the workload requires frequent memory access relative to computation, 
meaning the attainable performance is limited by memory bandwidth---a scenario referred to as \textit{memory-bound}. 
Conversely, a high arithmetic intensity implies that the operation performs many computations per byte transferred, 
and the performance bottleneck shifts to the GPU's compute capacity, known as a \textit{compute-bound} regime.

%{To provide an intuitive understanding of the relationship between arithmetic intensity and performance bottlenecks, we follow Kim \etal~\cite{16-arxiv-2025-minseo}, who adopt the roofline model~\cite{15-arxiv-2009-samue}, which relates a system’s attainable compute performance to its memory bandwidth and identifying whether a workload is limited by computation or memory access.} 
To understand the relationship between arithmetic intensity and performance, we adopt the roofline model \cite{15-arxiv-2009-samue} for our analysis, following the approach of Kim \etal \cite{16-arxiv-2025-minseo}.
This model relates a system's peak compute performance to its memory bandwidth, identifying whether a workload is compute-bound or memory-bound. 
In this framework, the arithmetic intensity ridge point  ($\text{Arithmetic Intensity}_{\text{ridge}}$) defines the boundary between the two regimes (compute-bound and memory-bound) and is calculated as follows:

\begin{equation}
    \text{Arithmetic Intensity}_{\text{ridge}} = \frac{\text{Peak FLOP Performance}}{\text{Peak Memory Bandwidth}} \; (\text{FLOPs/Byte}).
\end{equation}

According to the roofline model, workloads with arithmetic intensity below this ridge point are memory-bound regime, while those above it are compute-bound regime. %Hence, arithmetic intensity is a key indicator for analyzing and interpreting performance bottlenecks in model inference.
{
In the memory-bound regime, the attainable performance $P$ scales linearly with arithmetic intensity, with a slope equal to the peak sustainable memory bandwidth $B_{\text{mem}}$:
\[
P = B_{\text{mem}} \times \text{Arithmetic Intensity}.
\]
In contrast, in the compute-bound regime, the performance reaches the flat ceiling defined by the peak floating-point performance $P_{\text{max}}$.
}

% To visualize how arithmetic intensity interacts with hardware constraints, 
% we employ the \textbf{Roofline model}~\cite{Williams2009Roofline}.
% In this framework, the attainable performance is upper-bounded by two factors: 
% the system's \textbf{peak computational throughput} and its \textbf{peak memory bandwidth}. 
% The intersection of these two bounds defines the \textbf{ridge point}, 
% which represents the minimum arithmetic intensity required to fully utilize the compute capacity of a device. 
% Workloads with arithmetic intensity below this ridge are categorized as \textit{memory-bound}, 
% while those above it are \textit{compute-bound}.

% Intuitively, the ridge point reflects the overall balance of a hardware system:
% a higher ridge point implies that only workloads with extremely high arithmetic intensity can reach peak utilization, 
% whereas a lower ridge point indicates that even moderately intensive workloads can achieve near-optimal performance.

%% file: sections/evaluation_analysis.tex
\section{Evaluation and Analysis}
\label{sec:evaluation_analysis}

\subsection{RQ1: How do DLMs compare with AR Models in efficiency?}
\label{sec:rq1_effect}
For this question, we compare the efficiency of three types of models—DLM, AR, and block diffusion. We further investigate the factors influencing their efficiency and discuss how these models’ efficiency varies across different scenarios. Here, we consider two main factors, sequence length and batch size, which are critical for all these models.

% Figures
% \begin{figure}[htb]
%     \centering
%     \begin{subfigure}[b]{0.48\linewidth}
%         \centering
%         \includegraphics[width=\linewidth]{figures/figure_1.pdf}
%         \caption{}
%         \label{fig:scaling_a}
%     \end{subfigure}
%     \hfill
%     \begin{subfigure}[b]{0.48\linewidth}
%         \centering
%         \includegraphics[width=\linewidth]{figures/figure_2.pdf}
%         \caption{}
%         \label{fig:scaling_b}
%     \end{subfigure}
%     \caption{(a) and (b) show the throughput comparison across different generation lengths under the 0-shot and 5-shot settings, respectively.}
%     \label{fig:performance_throughput}
% \end{figure}

\begin{figure}[tb]
    \centering
    % ---------- (a) ----------
    \begin{subfigure}[b]{0.32\linewidth}
        \centering
        \includegraphics[width=\linewidth]{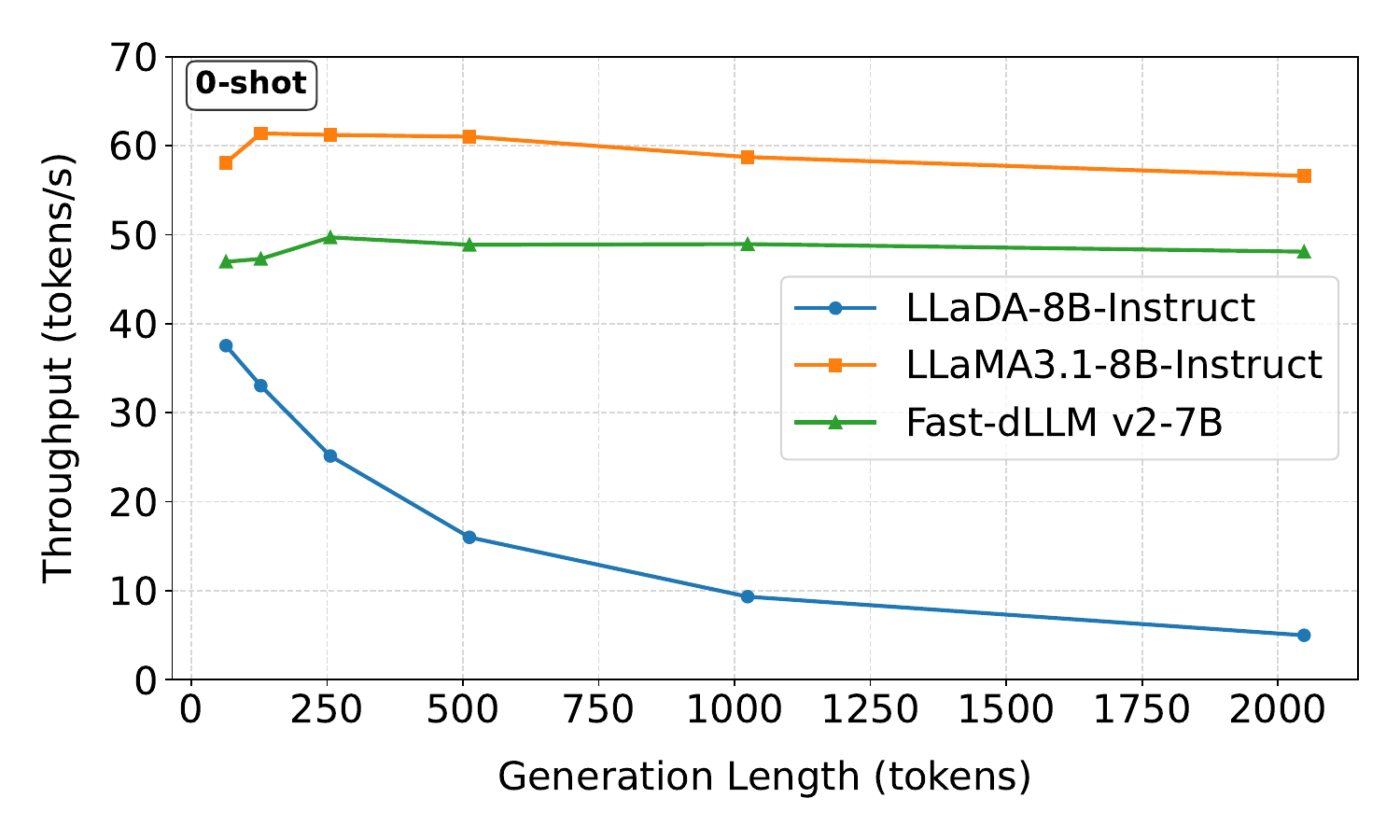}
        \caption{Throughput~(0-shot).}
        \label{fig:scaling_a}
    \end{subfigure}
    % \hfill
    \hspace{-6pt}
    % ---------- (b) ----------
    \begin{subfigure}[b]{0.32\linewidth}
        \centering
        \includegraphics[width=\linewidth]{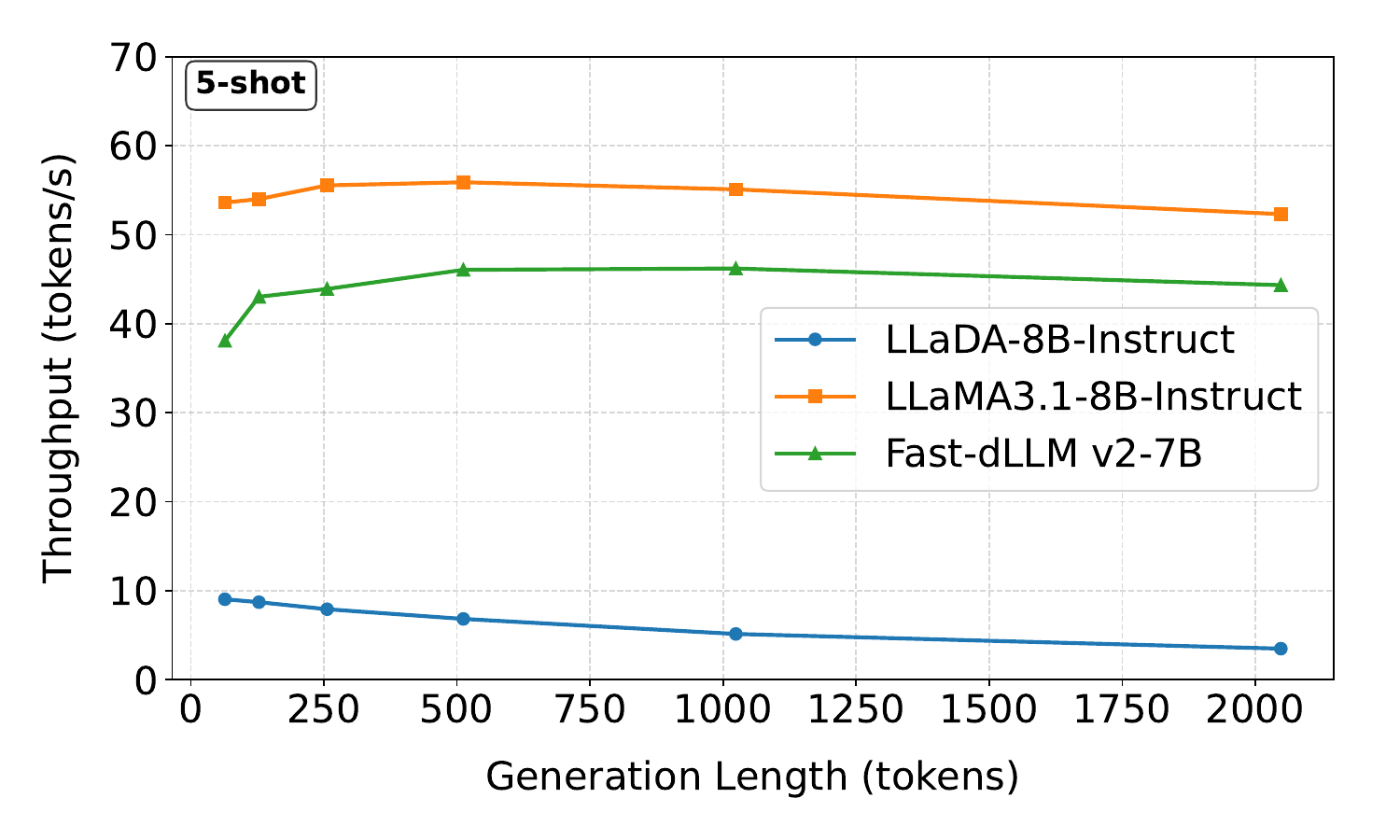}
        % \caption{5-shot setting.}
        \caption{Throughput~(5-shot).}
        \label{fig:scaling_b}
    \end{subfigure}
    % \hfill
    \hspace{-6pt}
    % ---------- (c) ----------
    \begin{subfigure}[b]{0.32\linewidth}
        \centering
        \includegraphics[width=\linewidth]{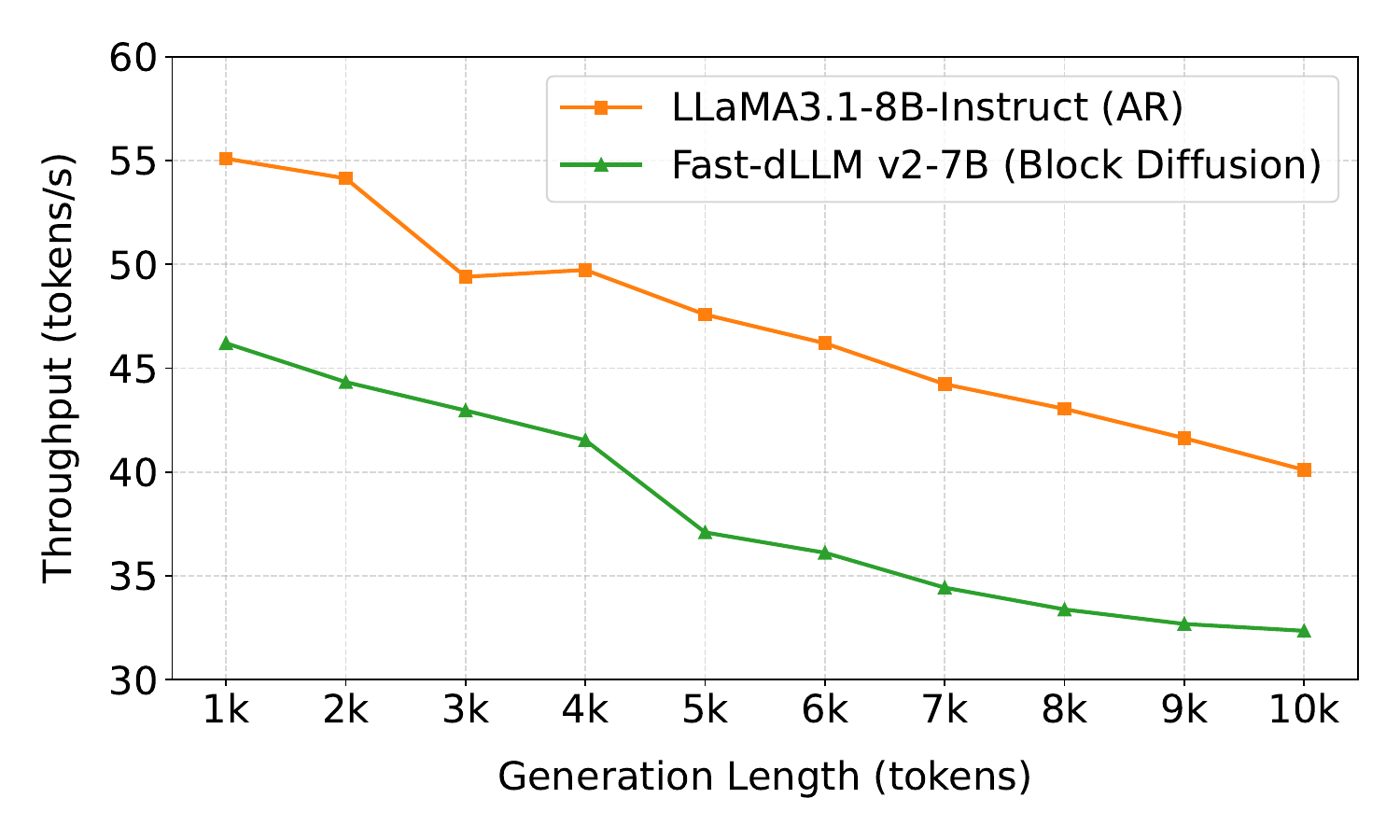}
        \caption{Throughput~(long generation).}
        \label{fig:scaling_c}
    \end{subfigure}
    % ---------- Caption ----------
    \caption{(a) and (b) show the throughput comparison across different generation lengths under the 0-shot and 5-shot settings, respectively, while (c) presents the throughput comparison between AR and block diffusion models under longer generation lengths.}
    \label{fig:rq1_len}
\end{figure}
% \vspace{-0.3cm}

\subsubsection{Effect of Sequence Length}
\label{sec:rq1_length_effect}

\paratitle{Experimental Settings.}
We select LLaDA-8B-Instruct, LLaMA3.1-8B-Instruct, and Fast-dLLM v2-7B as the representative models for DLM, AR and block diffusion. Subsequently, we fix the batch size as 1 and evaluate the effect of sequence lengths on model efficiency. Specifically, we compare the model performance from two aspects: prompt length and generation length. For convenience, we sample 100 examples from the GSM8K dataset for evaluation. We conduct both 0-shot and 5-shot experiments, corresponding to short and long prompt lengths, respectively. For each experiment setting, we evaluate different generation lengths, considering values of 64, 128, 256, 512, 1024, and 2048 tokens. Since AR cannot precisely control their effective generation length, we enforce them to generate an ``<eos>'' token until the generation exceeds the target length for fair comparison. We adopt a similar approach for block diffusion.

\paratitle{Main Results.}
We present the results of short and long prompt length in Figure~\ref{fig:rq1_len}.

First, across different generation lengths and prompt settings, the AR is faster than block diffusion, which in turn is faster than DLM. We can observe that regardless of the input and output lengths, the throughput of the DLM is significantly lower than that of AR and block diffusion with comparable size. We infer that DLM requires encoding the entire sequence when modeling each token, which greatly increases computational cost. Similar to AR, block diffusion is unidirectional, but its throughput decreases due to the diffusion-based modeling within each block.

Second, as the generation length increases, the throughput of DLM drops rapidly, while AR and block diffusion remain relatively stable within the 2K tokens. Under different prompt lengths, we can consistently observe that the throughput of the DLM drops rapidly as the generation length increases. In contrast, the throughput of the other two models remains roughly constant around a sequence length of 2K. As we extend the generation length further, their throughput decreases gradually, though at a much slower rate than the DLM.

Finally, increasing the prompt length leads to a decrease in throughput for all three models, with the DLM being the most affected. We compared the throughput variations of the three models under short and long prompt lengths. We can observe that the throughput of all three models decreases as the input length increases. Among them, AR and block diffusion show only a slight decline (around 10\%), and maintain relatively stable throughput across different input lengths. In contrast, the DLM exhibits a much larger drop, especially on shorter generation length. Specifically, the DLM’s throughput drops by about 75\% at a generation length of 64 and by around 50\% at 1024. We hypothesize that this behavior results from the quadratic computational complexity of DLM with respect to sequence length.

\begin{figure}[tb]
    \centering
    \begin{subfigure}[b]{0.48\linewidth}
        \centering
        \includegraphics[width=\linewidth]{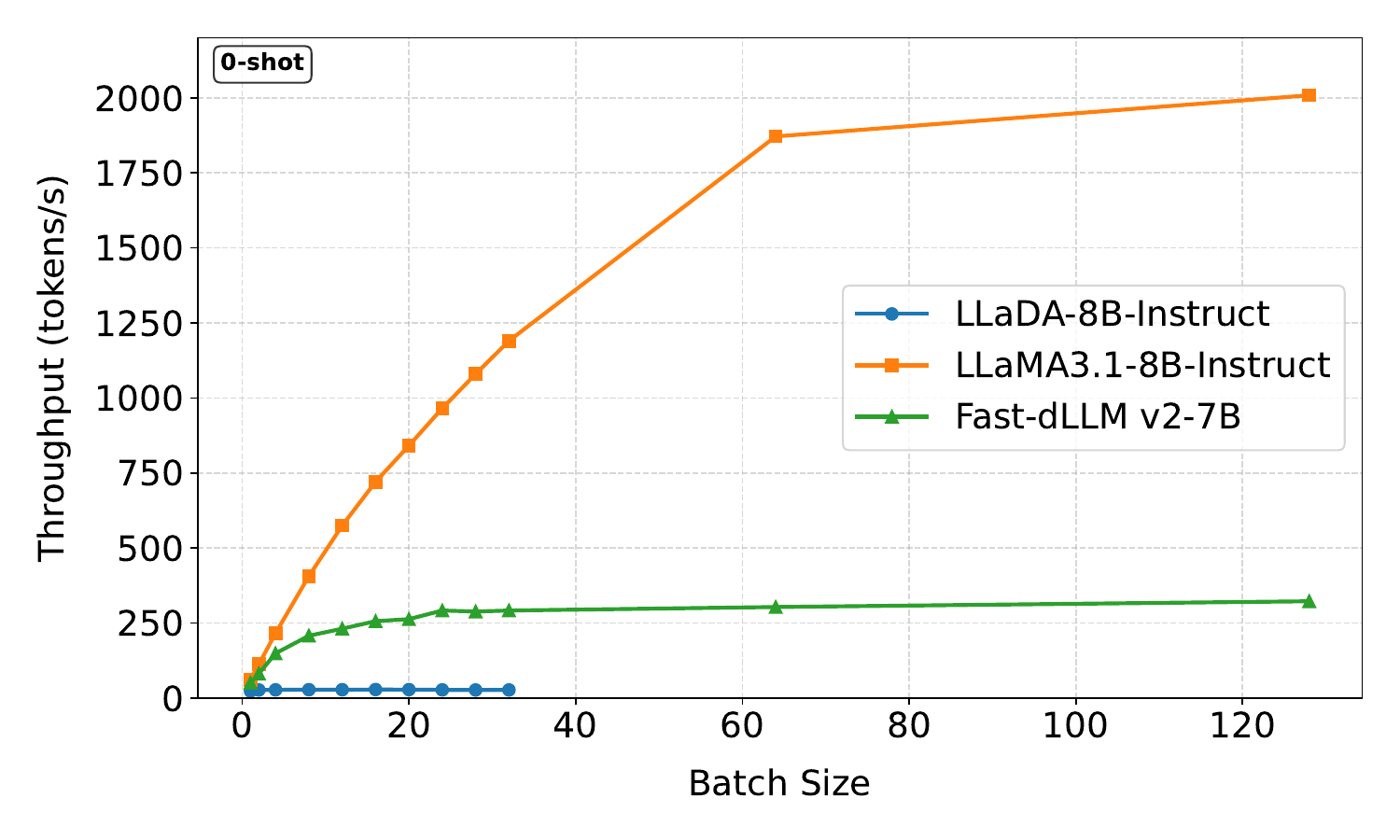}
        % \caption{Results under the 0-shot setting.}
        \caption{Throughput~(0-shot).}
        \label{fig:scaling_a}
    \end{subfigure}
    \hfill
    \begin{subfigure}[b]{0.48\linewidth}
        \centering
        \includegraphics[width=\linewidth]{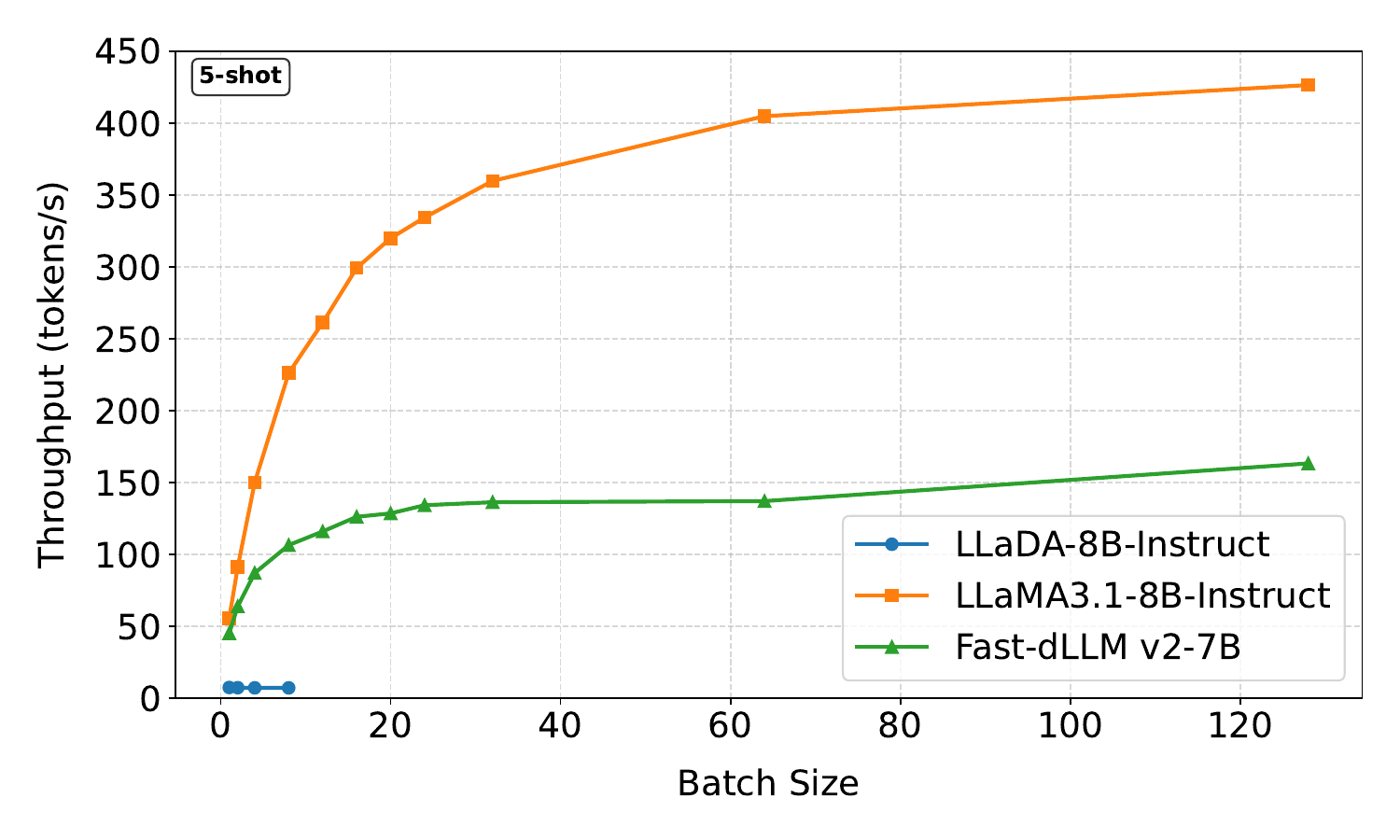}
        % \caption{Results under the 5-shot setting.}
        \caption{Throughput~(5-shot).}
        \label{fig:scaling_b}
    \end{subfigure}
    % \caption{(a) and (b) show the throughput comparison across different batch sizes under the 0-shot and 5-shot settings, respectively.}
    \caption{The throughput comparison across different batch sizes under the 0-shot~(a) and 5-shot~(b) settings, respectively.}
    \label{fig:rq1_bs}
\end{figure}

\subsubsection{Effect of Batch Size}
\paratitle{Experimental Settings.}
We choose the same model and data as the evaluation of sequence length. We fix the generation length as 256 tokens, and prompt length as 40 (0-shot) and 920 (5-shot) tokens. Subsequently, we evaluate the changes of throughput under different batch sizes, \ie 1, 2, 4, 8, 16, 20, 24, 32, and 64.

\paratitle{Main Results.}
We present the results of effect of batch size under the settings of short and long prompt lengths in Figure~\ref{fig:rq1_bs}.

First, consistent with the evaluation on sequence lengths in Section~\ref{sec:rq1_length_effect}, DLM remains slower than block diffusion, which in turn is slower than AR. Similar to the observations in analysis of sequence lengths, the DLM is also significantly slower than the other models.

Second, the throughput of DLM remains consistent across different batch sizes, but it eventually hits GPU memory limits at larger batch sizes. As the batch size increases, the throughput of the DLM is consistent, which may be owing to the compute-bound decoding. Additionally, the memory cost increases sharply, becoming out-of-memory with the batch sizes of 16 and 64 under the long and short prompt lengths.

Finally, as the batch size increases, the throughput of both AR and block diffusion grows steadily until it eventually stabilizes. Unlike DLM, the throughput of AR and block diffusion rises with larger batch sizes. However, after a certain point, the throughput of block diffusion plateaus, resembling the behavior of the DLM, whereas the AR model keeps increasing.

\begin{tcolorbox}[colback=blue!5!white,colframe=blue!55!black,width=1.0\textwidth,title={Summary}]
{
Based on the default implementation described in Section~\ref{sec-models}, the throughput of existing DLMs remains lower than that of AR and block diffusion models. The main findings are summarized as follows:

\vspace{0.15cm}
\textit{$\bullet$ For sequence length, throughput of DLM decreases sharply as generation length increases, while AR and block diffusion models remain stable up to around 2K tokens and then decline gradually.}

\vspace{0.15cm}
\textit{$\bullet$ For batch size, throughput of DLM stays nearly constant, whereas AR and block diffusion scale efficiently—rising rapidly at first and then gradually stabilizing as batch size increases.}
}
\end{tcolorbox}

\subsection{RQ2: How can we theoretically explain the variations in inference throughput?}
\label{sec:rq2_theory}

% 先说下这一段要做什么，他们做了什么，我们做了什么
% 我们在这里的目的是分析，在不同的类型模型中，吞吐量随着seq len 以及 bs的变化。
% 已经有一些工作，他们做了什么
To better understand the throughput variations observed in RQ1, we conduct a theoretical analysis of how inference throughput varies with critical factors (\ie sequence length and batch size) across different model architectures.
In a prior study, Kim \etal~\cite{16-arxiv-2025-minseo} adopt the roofline model~\cite{15-arxiv-2009-samue} to derive asymptotic formulations of FLOPs, MOPs, and arithmetic intensity for AR and DLMs, thereby revealing whether each model operates in a compute- or memory-bound regime. 
This analysis provides valuable intuition about hardware utilization.
% While this analysis provides valuable intuition about hardware utilization, it focuses primarily on arithmetic intensity rather than the end-to-end throughput.
Inspired by their work, we focus on the overall inference throughput and develop an extended analytical framework that explicitly connects model-side computational efficiency (FLOPs per token) with hardware-side computational performance (FLOPs per second). This formulation allows us to interpret throughput variations through both hardware and model-architecture perspectives, revealing how architectural design, sequence length, and batch size affect inference efficiency.

\subsubsection{Theoretical Analysis of Inference Throughput}
% To interpret the throughput variations observed in RQ1, we analyze how inference throughput is jointly affected by hardware attainable performance and model-side decoding design.
To formalize how inference throughput is influenced by hardware performance and model design, we decompose throughput into two components: (1) \textbf{hardware-side computational performance}, representing the amount of computation effectively executed per unit time (FLOPs/s) and determined by the workload’s arithmetic intensity and the roofline model; and (2) \textbf{model-side computational efficiency}, defined as the average computational cost required to generate a single token (FLOPs/token).
Formally, the inference throughput can be expressed as the ratio between these two components, as shown below:
% ---
% To analyze throughput quantitatively, we begin with its standard definition as the number of generated tokens per unit time, which depends on both the hardware capability and the model design.
% To obtain a measurable and comparable metric, we perform a simple transformation by introducing FLOPs into the formulation, yielding two interpretable terms:
% (1) \textbf{hardware-side computational performance}, represented by the achievable computation rate (FLOPs/s); and
% (2) \textbf{model-side computational workload}, defined as the average amount of computation required to generate one token (FLOPs/token).
% We summarize these two components and their corresponding analytical formulations in Table~\ref{tab:model_side_efficiency}, and next describe how each part can be approximately derived.}
% To interpret the throughput variations observed in RQ1, we introduce a unified analytical perspective that decomposes inference throughput into two components: (1) \textbf{hardware-side computational performance}, representing the amount of computation effectively executed per unit time (FLOPs/s) and determined by the roofline model and the workload’s arithmetic intensity; and (2) \textbf{model-side computational efficiency}, defined as the average computational cost required to generate a single token (FLOPs/token).
% Accordingly, the inference throughput can be formally expressed as:

\begin{equation}
\text{Throughput}=\frac{\text{Generated Tokens}}{\text{Generation Time}}\\=\frac{\text{FLOPs}/\text{Generation Time}}{\text{FLOPs}/\text{Generated Tokens}}=\frac{\text{FLOPs~per~second}}{\text{FLOPs ~per~token}}.
\end{equation}

% 这里吞吐的公式表述有问题，多加一些解释，对于分子分母的解释
% \paratitle{Hardware-Side Performance.} The roofline model characterizes the upper bound of attainable computational performance under given hardware constraints. And arithmetic intensity, defined as the ratio between computation and memory access, determines whether a workload is memory-bound (limited by bandwidth) or compute-bound (limited by the peak FLOP performance). When arithmetic intensity is below the ridge point, the effective compute is limited by memory bandwidth; when above, it reaches the compute bound determined by the peak FLOPs.

\begin{figure}[t]
    \centering
    % ---------- (a) ----------
    \begin{subfigure}[b]{0.32\linewidth}
        \centering
        \includegraphics[width=\linewidth]{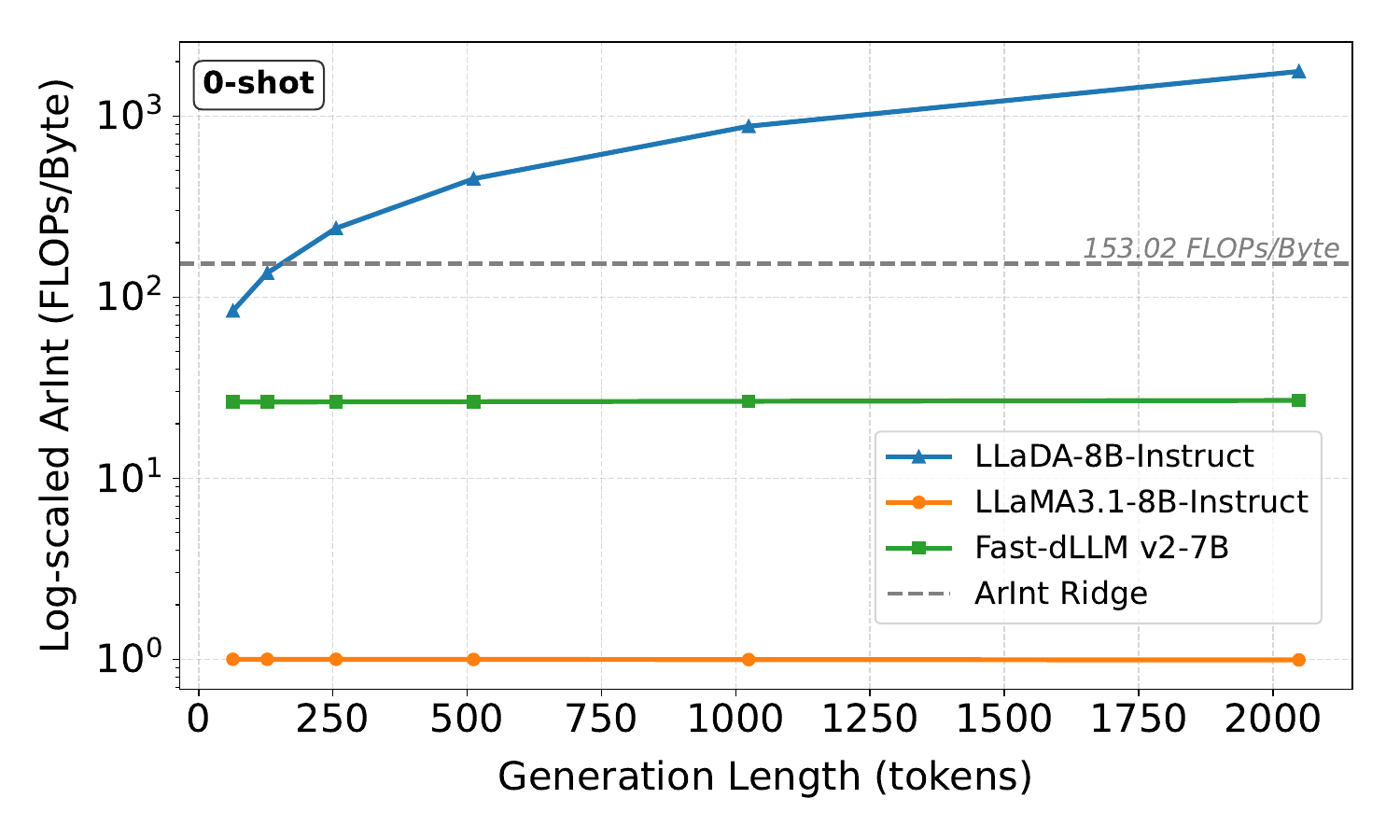}
        \caption{ArInt~(0-shot).}
        \label{fig:scaling_a}
    \end{subfigure}
    % \hfill
    \hspace{-6pt}
    % ---------- (b) ----------
    \begin{subfigure}[b]{0.32\linewidth}
        \centering
        \includegraphics[width=\linewidth]{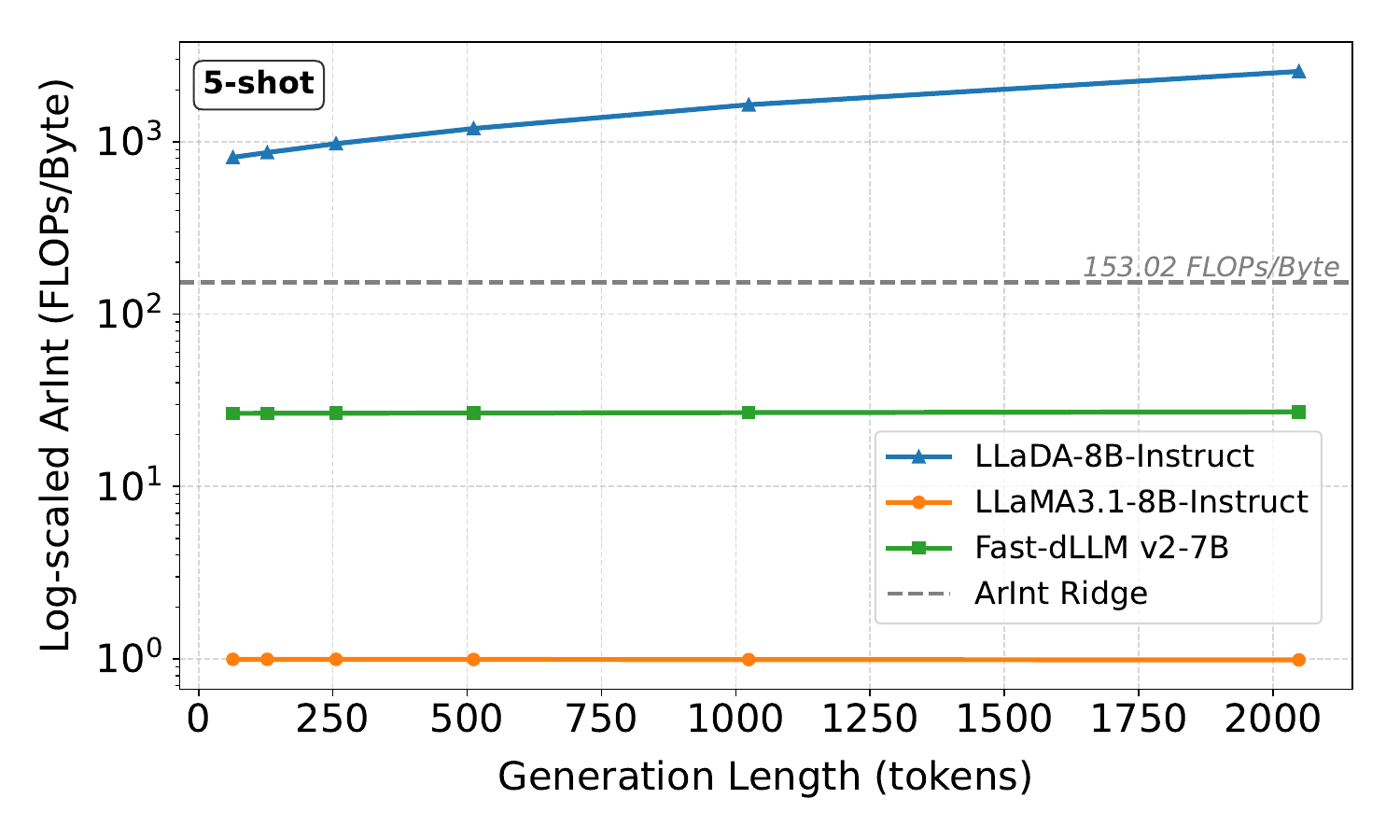}
        % \caption{5-shot setting.}
        \caption{ArInt~(5-shot).}
        \label{fig:scaling_b}
    \end{subfigure}
    % \hfill
    \hspace{-6pt}
    % ---------- (c) ----------
    \begin{subfigure}[b]{0.32\linewidth}
        \centering
        \includegraphics[width=\linewidth]{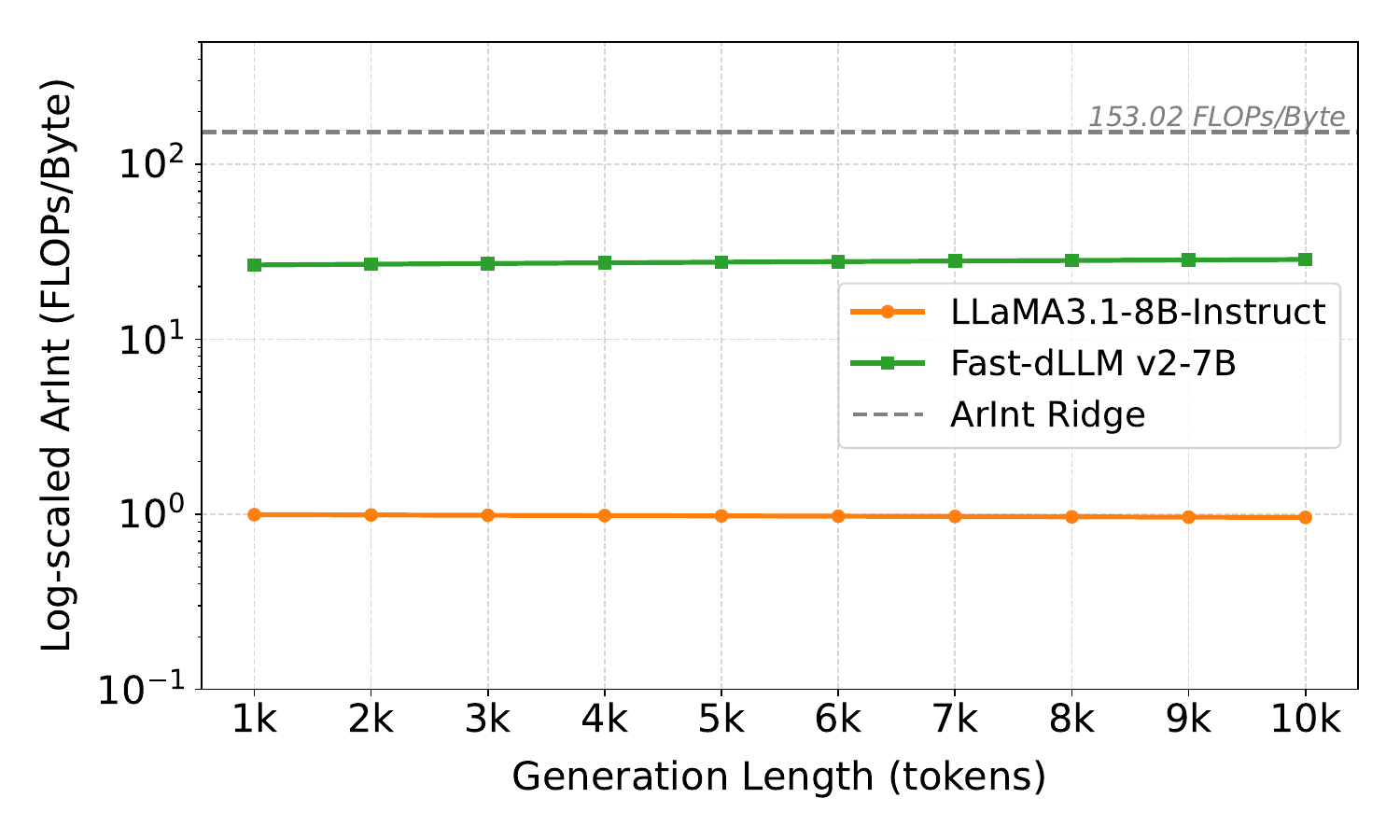}
        \caption{ArInt~(long generation).}
        \label{fig:scaling_c}
    \end{subfigure}
    % ---------- Caption ----------
    \caption{(a) and (b) show the arithmetic intensity comparison across different generation lengths under the 0-shot and 5-shot settings, respectively, while (c) presents the arithmetic intensity comparison between AR and block diffusion models under longer generation lengths.}
    \label{fig:rq2_len}
\end{figure}

\begin{figure}[t]
    \centering
    \begin{subfigure}[b]{0.48\linewidth}
        \centering
        \includegraphics[width=\linewidth]{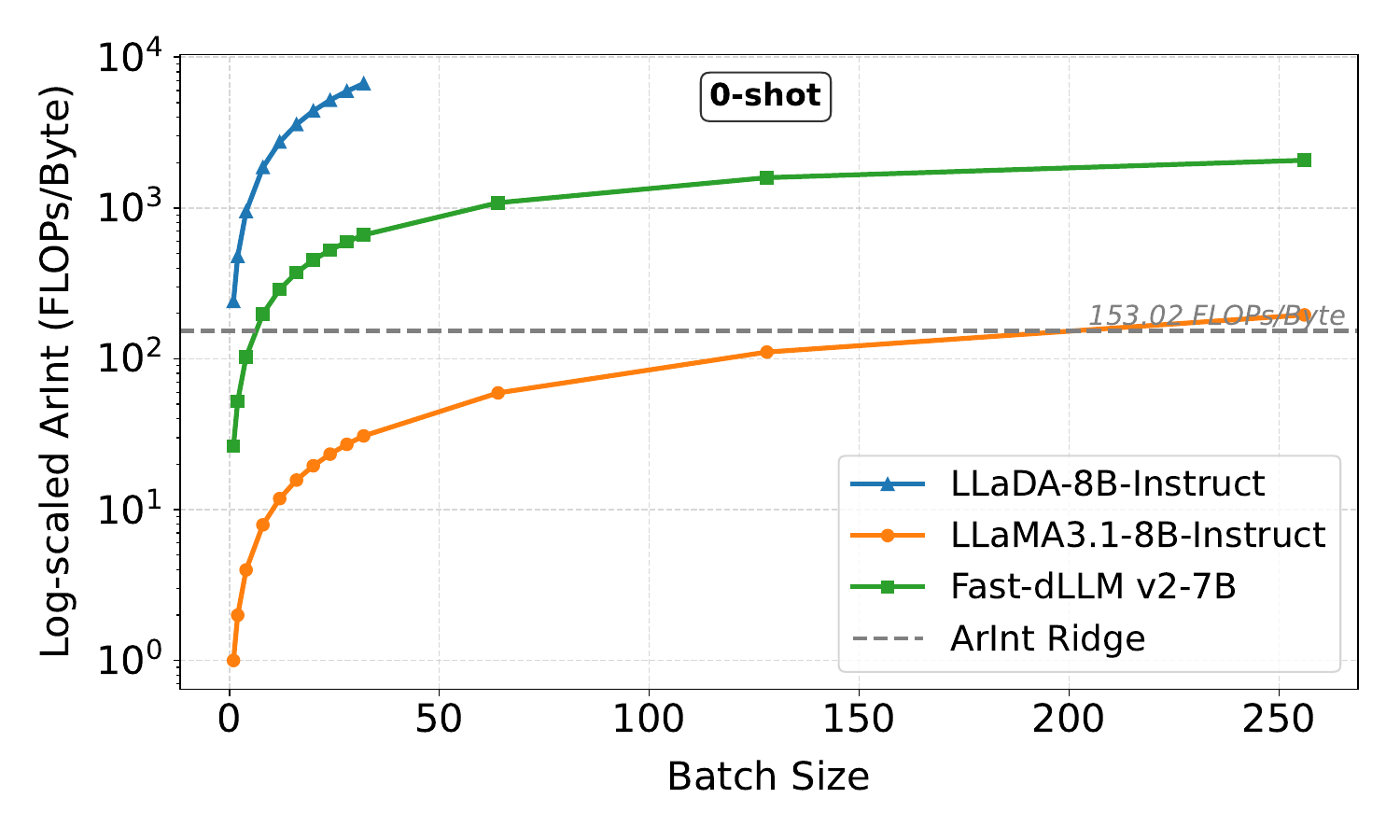}
        % \caption{Results under the 0-shot setting.}
        \caption{ArInt~(0-shot).}
        \label{fig:scaling_a}
    \end{subfigure}
    \hfill
    \begin{subfigure}[b]{0.48\linewidth}
        \centering
        \includegraphics[width=\linewidth]{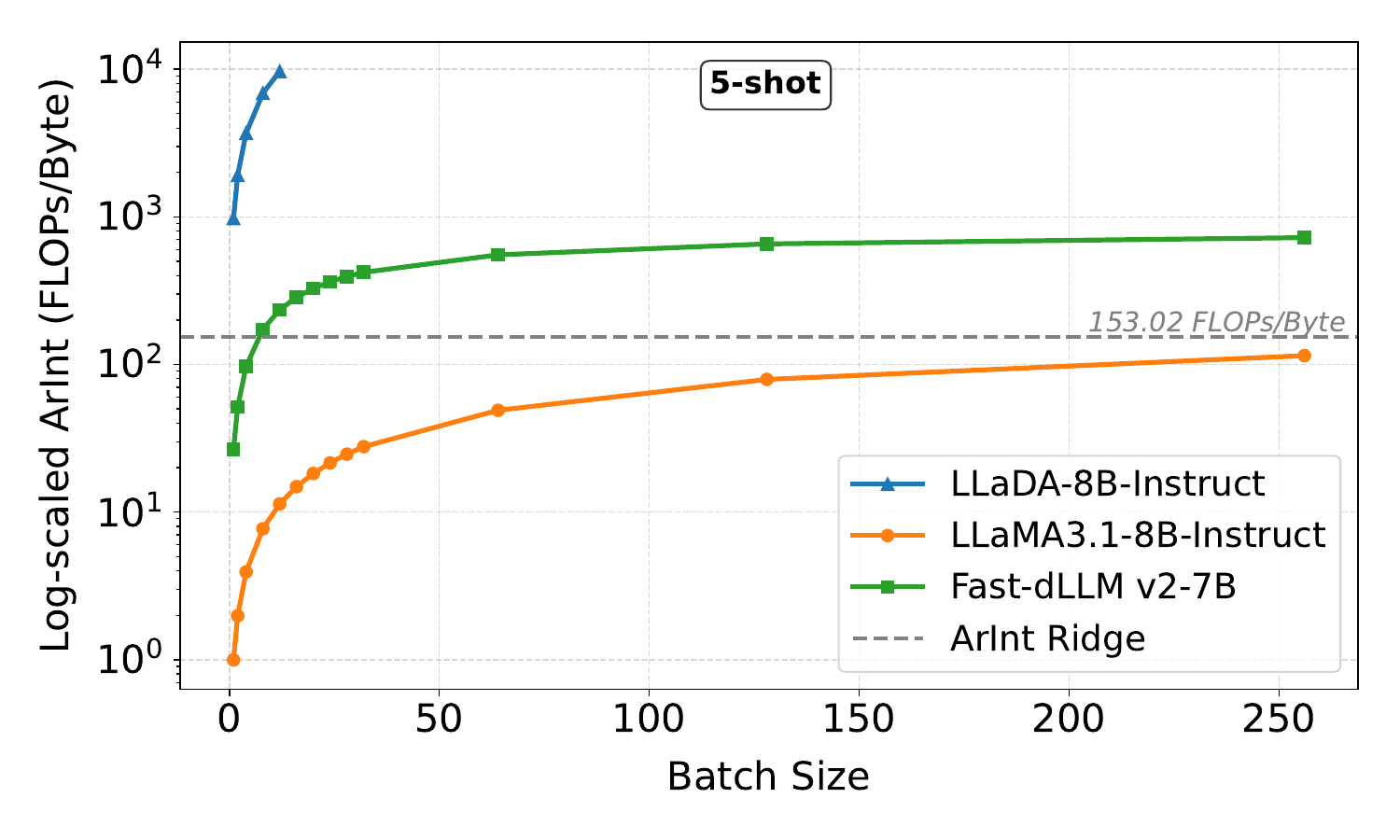}
        % \caption{Results under the 5-shot setting.}
        \caption{ArInt~(5-shot).}
        \label{fig:scaling_b}
    \end{subfigure}
    % \caption{(a) and (b) show the throughput comparison across different batch sizes under the 0-shot and 5-shot settings, respectively.}
    \caption{The arithmetic intensity comparison across different batch sizes under the 0-shot~(a) and 5-shot~(b) settings, respectively.}
    \label{fig:rq2_bs}
\end{figure}

\paratitle{Hardware-Side Performance.} 
% 为了拿到hardware attainable performance，即FLOPs per second，根据4.2.4的理论解释，我们需要首先获取模型的算术强度，之后经过roofline模型的分析来估算实际的hardware attainable performance. 这里我们参考kim等人的asymptotic formulation，计算了不同模型在memory- and compute- bound下的attainable performance，见Table 1中的两行FLOPs per second.
% 加个转折，就说specifically，我们首先基于他们的工作，结合roofline给出了简化版本的FLOPs per second的渐近表达式
% 在此基础上，为了有效区分在不同settings下，模型的workload到底如何以及其最终的FLOPs per second，我们给出了RQ1中所有实验配置对应的算术密度和我们硬件的算术强度ridge. 如图3和图4所示，图中低于ArInt ridge线的实验配置可以近似认为是memory-bound，超过的实验配置可以近似认为是compute-bound.
% 综合上述的分析，我们可以清晰给出在具体配置下模型的算术密度，进一步通过roofline model分析得到最终的hardware attainable performance，即FLOPs per second.
% 根据4.2.4的叙述，硬件侧的性能由
To estimate the hardware-side attainable performance, we first compute the arithmetic intensity of each model and then apply the roofline model to obtain its attainable FLOPs per second.
Specifically, based on the formulations proposed by Kim \etal~\cite{16-arxiv-2025-minseo}, we derive simplified asymptotic expressions for FLOPs per second under both memory- and compute-bound conditions, as summarized in Table~\ref{tab:hardware_model_efficiency} (rows of FLOPs per second).
% 之后，我们进一步更细粒度估计了RQ中实践实验设置下的算术密度估计值，如图3图4所示
Building on the formulation in Table~\ref{tab:hardware_model_efficiency}, we further perform detailed approximations of arithmetic intensity under the practical experimental settings used in RQ1, with the corresponding formulas provided in Appendix~\ref{app:arint_est}.
Figure~\ref{fig:rq2_len} and Figure~\ref{fig:rq2_bs} present these estimated values together with the hardware-specific arithmetic intensity ridge, which separates the memory-bound and compute-bound regimes. Configurations below the ridge are approximately memory-bound, whereas those above are compute-bound. This analysis allows us to infer the attainable computational performance for each setting, providing the basis for the subsequent throughput interpretation.
\begin{table}[t]
\small
\centering
\renewcommand{\arraystretch}{2.5}
\setlength{\tabcolsep}{2pt}
\caption{{
Analysis of FLOPs per second and FLOPs per token
for AR, block diffusion, and DLMs.
The gray rows are derived based on the asymptotic analysis proposed by Kim \etal~\cite{16-arxiv-2025-minseo}, following their approximate formulations and notations.
Here, $B_\text{mem}$ denotes the peak sustainable memory bandwidth, $P_{\text{max}}$ is the peak floating-point performance, $L = L_p + L_g$ (the total length of prompt and generated response), 
$B$ is batch size, $d$ is hidden dimension, 
$K$ is the number of diffusion steps, and $G$ is the block size.}}
\vspace{0.5cm}
\label{tab:hardware_model_efficiency}
\begin{tabular}{lccc}
\toprule
 & \textbf{AR} & \textbf{Block Diffusion} & \textbf{DLM} \\
\midrule

\makecell[l]{\textbf{FLOPs per second}\\(memory-bound)} &
$\displaystyle % \approx

\cellcolor[HTML]{F7F7F7}
\begin{cases}
\mathcal{O}(B_\text{mem}B),\ L \ll d \\
\mathcal{O}(B_\text{mem}),\ L \gg d
\end{cases}$ &
$\displaystyle % \approx
\cellcolor[HTML]{F7F7F7}
\begin{cases}
\mathcal{O}(B_\text{mem} B G), \ L \ll d \\
\mathcal{O}(B_\text{mem} G), \ L \gg d
\end{cases}$ &
$\displaystyle % \approx
\cellcolor[HTML]{F7F7F7}
\begin{cases}
\mathcal{O}(B_\text{mem}B L), \ L \ll d \\
\mathcal{O}(B_\text{mem}L), \ L \gg d
\end{cases}$ \\

\makecell[l]{\textbf{FLOPs per second}\\(compute-bound)} &
$P_{\text{max}}$ &
$P_{\text{max}}$ &
$P_{\text{max}}$ \\

\textbf{Generated Tokens} &
$B L_g$ &
$B L_g \approx B K$ &
$B L_g \approx B K$ \\

\makecell[l]{\textbf{FLOPs per token}} &
\cellcolor[HTML]{F7F7F7}
$\mathcal{O}(d^{2}) + \mathcal{O}(L d)$ &
\cellcolor[HTML]{F7F7F7}
$\mathcal{O}(G d^{2}) + \mathcal{O}(G L d)$ &
\cellcolor[HTML]{F7F7F7}
$\mathcal{O}(L d^{2}) + \mathcal{O}(L^{2} d)$ \\

\bottomrule
\end{tabular}
\end{table}

\paratitle{Model-Side Efficiency.} As shown in Equation~\ref{eq:flops_per_token}, the average FLOPs per token is a statistical measure of the model’s computational efficiency. It is obtained by dividing the model’s theoretical asymptotic FLOPs—representing the total floating-point operations required for the entire generation process—by the total number of generated tokens. The results of the three architectural model types are shown in Table~\ref{tab:hardware_model_efficiency}.

\begin{equation}
\text{FLOPs/token} = \frac{\text{Total FLOPs for decoding}}{\text{Generated Tokens}}.
\label{eq:flops_per_token}
\end{equation}

\subsubsection{Analysis of AR Models}
In the decoding phase of AR models, the model generates only one token at each step, incurring per-token FLOPs of $\mathcal{O}(Bd^2)+\mathcal{O}(BLd)$ and MOPs of $\mathcal{O}(d^2)+\mathcal{O}(BLd)$ due to Key-Value (KV) cache reads. 
{According to Figure~\ref{fig:rq2_len},}
% roofline model, when the batch size is one, 
at the batch size of one, the arithmetic intensity of the AR decoding process is the lowest, making it a memory-bound workload. Consequently, its achievable throughput is mainly limited by memory bandwidth and can be asymptotically expressed as:

\begin{equation}
\text{Throughput} \approx\begin{cases}\dfrac{\mathcal{O}(B_{\text{mem}} B)}{\mathcal{O}(d^2)+\mathcal{O}(Ld)} = \dfrac{\mathcal{O}(B)}{\mathcal{O}(d^2)+\mathcal{O}(Ld)}, & L \ll d, \\[10pt]\dfrac{\mathcal{O}(B_{\text{mem}})}{\mathcal{O}(d^2)+\mathcal{O}(Ld)} = \dfrac{\mathcal{O}(1)}{\mathcal{O}(d^2)+\mathcal{O}(Ld)}, & L \gg d.\end{cases}
\label{eq:ar_throughput}
\end{equation}

% \textcolor{blue}{This formulation serves as the analytical basis for the subsequent analysis of how throughput varies with sequence length and batch size.}

{This formulation is applied in the analysis of generation length (with the batch size fixed at $1$) and batch size (under the condition $L \ll d$).}

\paratitle{Effect of Generation Length.} The throughput formula can be derived from Equation~\ref{eq:ar_throughput}:

\begin{equation}
\text{Throughput} \approx\begin{cases}\dfrac{\mathcal{O}(1)}{\mathcal{O}(d^2)}, & L \ll d, \\[10pt]\dfrac{\mathcal{O}(1)}{\mathcal{O}(Ld)}, & L \gg d.\end{cases}
\end{equation}

When the generation length is short, both the numerator and denominator in the throughput expression remain nearly constant (batch size fixed at 1). The denominator is dominated by the $\mathcal{O}(d^2)$ term, while the $\mathcal{O}(Ld)$ component is negligible. Consequently, the throughput remains stable with respect to sequence length. When the generation length is long, however, the $\mathcal{O}(Ld)$ term becomes dominant, so as $L$ increases, overall throughput decreases.

\paratitle{Effect of Batch Size.} In our evaluation setting (with generation length fixed at 256 tokens and relatively short prompts), the condition $L \ll d$ holds. In this regime, the denominator remains constant while the numerator scales linearly with batch size $B$ (because of the arithmetic intensity is $\mathcal{O}(B)$ ), leading to nearly linear throughput growth until the system reaches the compute roof. Once the arithmetic intensity reaches the compute roof, throughput remains constant as the workload becomes compute-bound.

\subsubsection{Analysis of Diffusion Models}
In decoding, DLMs perform denoising steps on the entire sequence of length $L=L_p+L_g$. The per-step computational cost scales as FLOPs $\mathcal{O}(BLd^2)+\mathcal{O}(BL^2d)$, with MOPs scaling as $\mathcal{O}(d^2)+\mathcal{O}(BLd)$ due to reading/writing activations.

Therefore, the arithmetic intensity satisfies:

\begin{equation}
\text{ArInt} \approx\begin{cases}\mathcal{O}(B L), & L \ll d,\\[6pt]\mathcal{O}(L),   & L \gg d.\end{cases}
\label{equ:dlm_arint}
\end{equation}

\paratitle{Effect of Generation Length.} For very short $L$ , the workload is still memory-bound. The throughput can be derived from Equation~\ref{equ:dlm_arint}:

% \begin{equation}
% \text{Throughput} \approx\frac{\mathcal{O}(BL)}{\mathcal{O}(BLd^2)+\mathcal{O}(BL^2d)}= \frac{\mathcal{O}(L)}{\mathcal{O}(Ld^2)+\mathcal{O}(L^2d)} = \frac{\mathcal{O}(1)}{\mathcal{O}(d^2)+\mathcal{O}(Ld)}.
% \end{equation}
\begin{equation}
\text{Throughput} \approx\frac{\mathcal{O}(B_{\text{mem}}BL)}{\mathcal{O}(BLd^2)+\mathcal{O}(BL^2d)}= \frac{\mathcal{O}(1)}{\mathcal{O}(d^2)+\mathcal{O}(Ld)}.
\end{equation}

Since $L \ll d$, the $\mathcal{O}(d^2)$ term dominates, so the throughput is roughly stable with respect to $L$.
% In most practical settings, the DLM is compute-bound. 
{However, in our experiments shown in Figure~\ref{fig:rq2_len}, the arithmetic intensity of the DLM quickly exceeds the ridge point, indicating that it soon operates in the compute-bound regime.}
In this case, the throughput is:

\begin{equation}
\text{Throughput} \approx \frac{P_\text{max}}{\mathcal{O}(BLd^2)+\mathcal{O}(BL^2d)}=\frac{P_\text{max}}{\mathcal{O}(Ld^2)+\mathcal{O}(L^2d)},
\end{equation}

where $P_{\text{max}}$ represents the peak floating-point performance (Peak FLOPs).
Here the numerator stays roughly constant, so as $L$ increases, the overall throughput decreases.

\paratitle{Effect of Batch Size.} 
% In our experiments, 
{As shown in Figure~\ref{fig:rq2_bs}}, the DLM remains compute-bound (with 
$\text{ArInt} \approx \mathcal{O}(B L)$
). Consequently, the throughput stays roughly flat as batch size $B$ increases, until it reaches the GPU memory limit. This trend is consistent with the observations in RQ1.

\subsubsection{Analysis of Block Diffusion Models}
In the decoding phase of block diffusion models, the sequence is partitioned into blocks of size G. Tokens within each block are generated in parallel via a diffusion process, incurring FLOPs of $\mathcal{O}(BGd^2)+\mathcal{O}(BGLd)$. Despite the inter-block parallelism, the block-level progression follows an autoregressive scheme, benefiting from KV cache reuse. 
% The resulting arithmetic intensity is inherently higher than that of standard AR models because the expensive KV cache reads are amortized across the parallel block generation, its throughput can be formulated as:
The resulting arithmetic intensity, {as shown in Figure~\ref{fig:rq2_len} and Figure~\ref{fig:rq2_bs}}, is inherently higher than that of standard AR models because the expensive KV cache reads are amortized across the parallel block generation. Consequently, its throughput can be formulated as:

\begin{equation}
\text{Throughput} \approx\begin{cases}\dfrac{\mathcal{O}(B_{\text{mem}}BG)}{\mathcal{O}(Gd^2)+\mathcal{O}(GLd)} = \dfrac{\mathcal{O}(BG)}{\mathcal{O}(Gd^2)+\mathcal{O}(GLd)}, & L \ll d, \\[10pt]\dfrac{\mathcal{O}(B_{\text{mem}}G)}{\mathcal{O}(Gd^2)+\mathcal{O}(GLd)} = \dfrac{\mathcal{O}(G)}{\mathcal{O}(Gd^2)+\mathcal{O}(GLd)}, & L \gg d.\end{cases}
\end{equation}

\paratitle{Effect of Generation Length.} Similar to AR, the block diffusion model is memory-bound when the batch size is 1. 
{After simplification, its throughput is the same as that of the AR model, and thus follows the same trend: it remains roughly constant for short $L$, where the computation is dominated by $\mathcal{O}(d^2)$~(Figure~\ref{fig:rq1_len}a-b) and decreasing as $L$ grows, where the complexity is dominated by $\mathcal{O}(Ld)$~(Figure~\ref{fig:rq1_len}c).}

\paratitle{Effect of Batch Size.} In our evaluation setting (with generation length fixed at 256 tokens and relatively short prompts), the condition {$L \ll d$} holds. For block diffusion models, the arithmetic intensity is $\mathcal{O}(BG)$, which is higher than AR under the same conditions. As a result, block diffusion reaches the compute-bound regime at a smaller batch size than AR. As batch size increases, the throughput of block diffusion models behaves much like AR: throughput grows almost linearly with $B$ until it hits the compute roof~{(Figure~\ref{fig:rq1_bs}a)}; beyond that point, throughput stays steady as the workload becomes compute-bound~{(Figure~\ref{fig:rq1_bs}b)}.

% summary of rq2
% \begin{promptbox}[Summary]{blue!60!black}
% We conduct a theoretical study to investigate the inference throughput bottlenecks across different language model architectures, with a particular focus on the computational characteristics of DLMs. Our key findings are as follows:

% $\bullet$ The higher arithmetic intensity of DLMs enables more effective utilization of GPU computational resources, resulting in higher FLOP throughput than AR models of comparable scale.

% $\bullet$ The average computational cost per generated token (FLOPs/token) in DLMs increases rapidly with sequence length, causing their overall inference throughput to remain lower than that of AR models in most settings.

% $\bullet$ Improving DLM efficiency requires reducing FLOPs per token while maintaining high arithmetic intensity to sustain strong computational utilization.
% \end{promptbox}
\begin{tcolorbox}[colback=blue!5!white,colframe=blue!55!black,width=1.0\textwidth,title={Summary}]
{
We conduct a theoretical study to investigate the inference throughput bottlenecks across different language model architectures, with a particular focus on the computational characteristics of DLMs. Our key findings are as follows:

\vspace{0.15cm}
\textit{$\bullet$ The higher arithmetic intensity of DLMs enables more effective utilization of GPU computational resources, resulting in higher FLOPs than AR models of comparable scale.}

\vspace{0.15cm}
\textit{$\bullet$ The average computational cost per generated token (FLOPs/token) in DLMs increases rapidly with sequence length, causing their overall inference throughput to remain lower than that of AR models in most settings.}

\vspace{0.15cm}
\textit{$\bullet$ Improving DLM efficiency requires reducing FLOPs per token while maintaining high arithmetic intensity to sustain strong computational utilization.
}}
\end{tcolorbox}

\subsection{RQ3: How do acceleration strategies benefit DLMs in varied conditions?}

In this part, we continue to analyze the effects and influencing factors of different acceleration strategies for DLM and block diffusion, which can be broadly categorized into two types: reducing the computational cost per denoising step (\eg ``+dual cache''), and decreasing the total number of denoising steps required for generation (\eg ``+parallel''). To ensure a fair and consistent evaluation, we follow the same experimental setup as in RQ1 and further examine how current acceleration techniques perform across a broader range of reasoning tasks in terms of inference efficiency. These analyses focus on the two primary external factors (\ie sequence length and batch size), which strongly influence inference efficiency, while keeping other variables fixed for controlled comparison. This section provides an analysis of the efficiency patterns of different acceleration strategies under varying sequence lengths and batch sizes.

\subsubsection{Effect of sequence length}

\begin{figure}[t]
    \centering
    \begin{subfigure}[b]{0.48\linewidth}
        \centering
        \includegraphics[width=\linewidth]{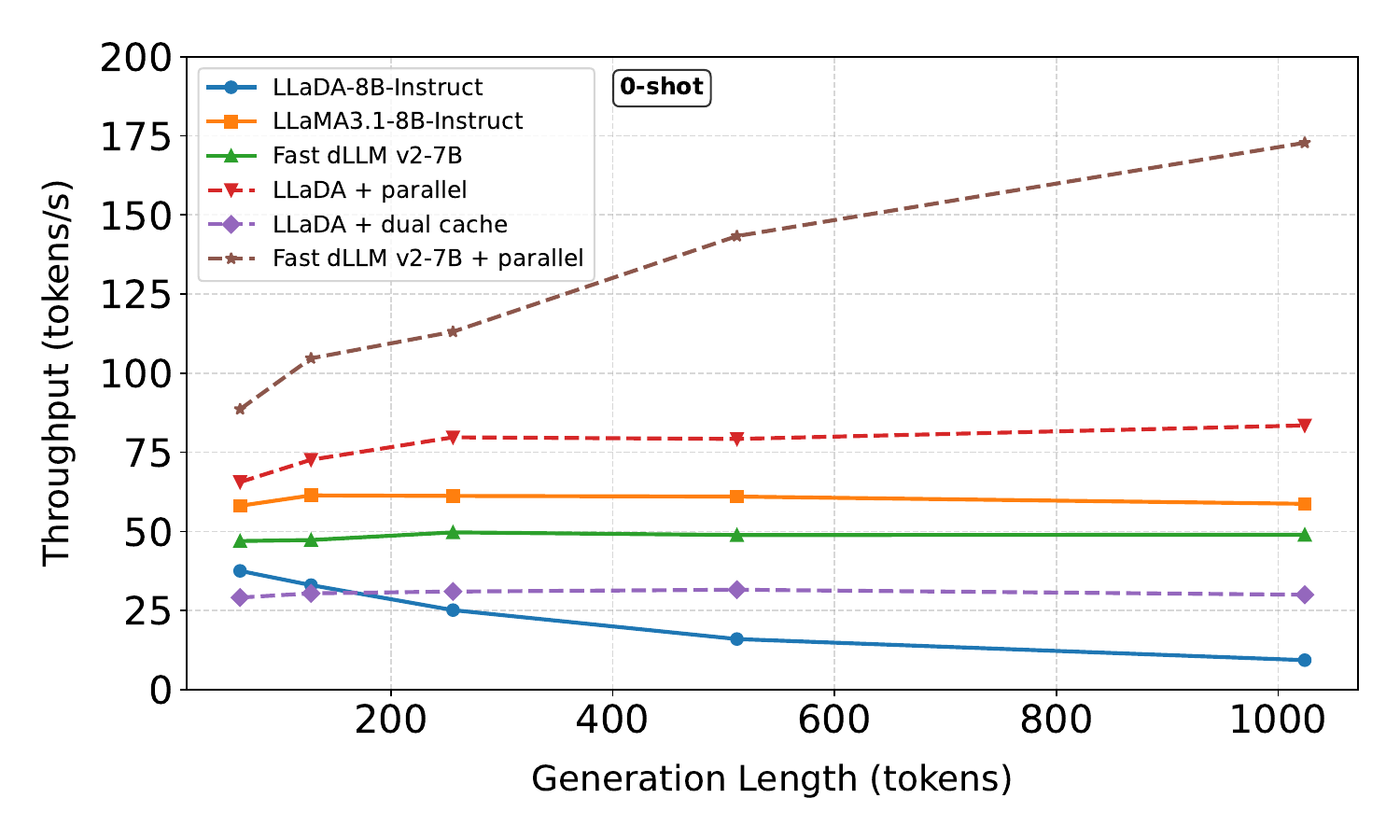}
        % \caption{Results under the 0-shot setting.}
        \caption{Throughput~(0-shot).}
        \label{fig:scaling_a}
    \end{subfigure}
    \hfill
    \begin{subfigure}[b]{0.48\linewidth}
        \centering
        \includegraphics[width=\linewidth]{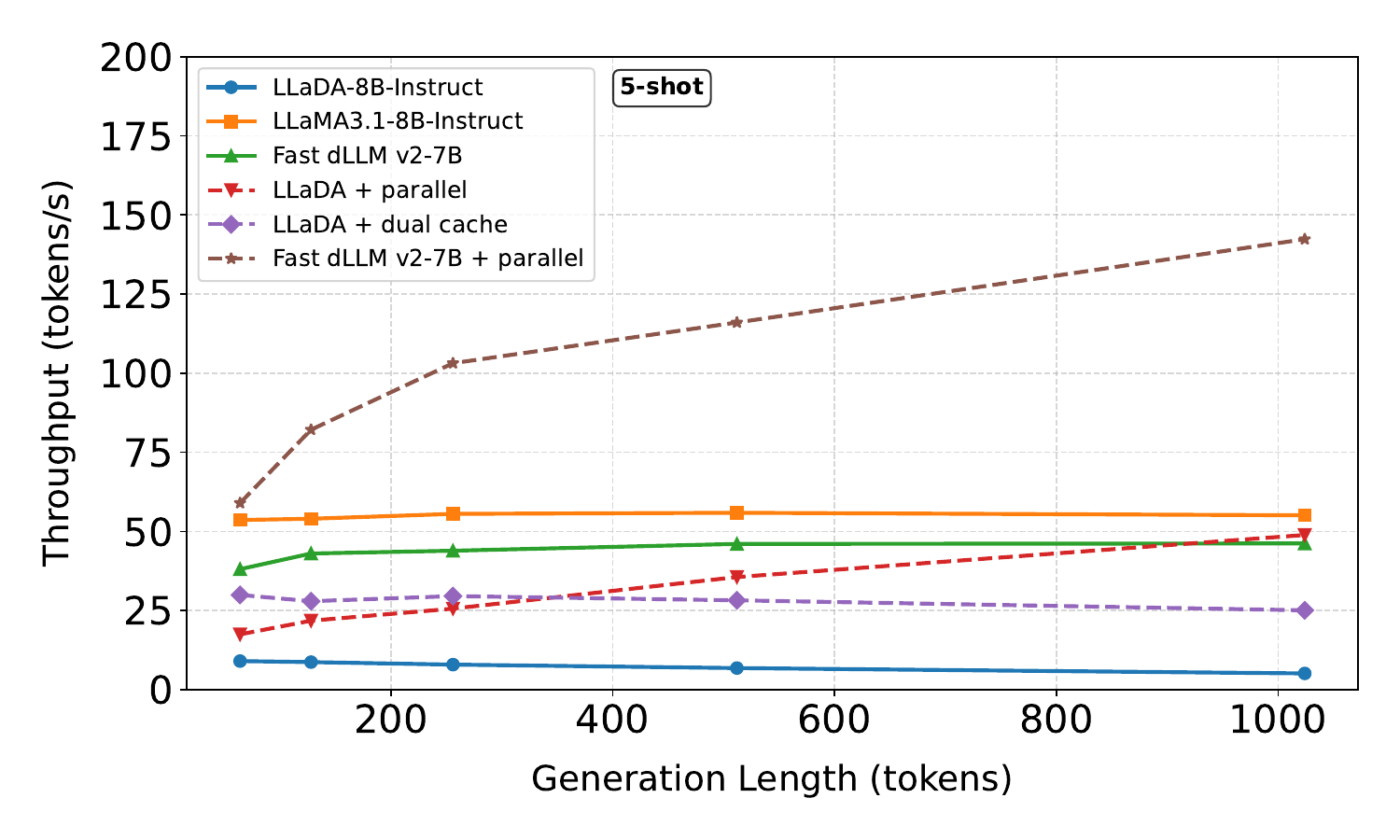}
        % \caption{Results under the 5-shot setting.}
        \caption{Throughput~(5-shot).}
        \label{fig:scaling_b}
    \end{subfigure}
    \caption{(a) and (b) show the throughput comparison across different generation lengths under the 0-shot and 5-shot settings, respectively.}
    \label{fig:rq3_len}
\end{figure}
% \vspace{-0.1cm}

\paratitle{Experimental Settings.}
We follow the same model and dataset configurations as in RQ1, with the only difference being the incorporation of specific acceleration strategies. Specifically, we evaluate the two representative strategies, ``+dual cache'' and ``+parallel''. For DLMs, both variants are implemented, while for block diffusion models, only the ``+parallel'' strategy is applicable due to their partially autoregressive design.

\paratitle{Main Results.}
The results are shown in the Figure~\ref{fig:rq3_len}.

For block diffusion models, the ``+parallel'' strategy is highly effective when evaluated with a batch size of 1, consistently achieving the fastest decoding performance across all sequence lengths and even outperforming the AR baseline. We observe that this acceleration effect further amplifies as sequence length increases, reaching up to \textbf{3.1×} speedup over the original configuration (\ie 142.33 vs. 46.20 tokens/s). This improvement stems from the model’s ability to generate multiple high-confidence tokens simultaneously within each forward pass, effectively reducing the number of decoding steps and increasing overall throughput. Consequently, the effectiveness of different parallel strategies can be evaluated by their degree of parallel generation, which can be quantified by the number of tokens per forward step (TPF)~\cite{17-arxiv-2025-yuxin}. This metric directly reflects how efficiently a model converts computational intensity into output throughput, providing a unified lens for analyzing diffusion-based acceleration methods.

For LLaDA, both acceleration strategies are effective when the batch size is 1, but their gains eventually saturate as the generation length grows. Specifically, the ``+parallel'' variant generally delivers higher throughput, except under long prompts (5-shot) with short outputs (generation length less than 256), where ``+dual cache'' performs better. In the remaining sequence-length regimes, ``+parallel'' attains the best efficiency—rising to a peak (\eg around 80 tokens/s at prefilling length 256 in the 0-shot setting, (a) in Figure~\ref{fig:rq3_len}) and then saturating; it can even surpass the AR baseline at moderate lengths. While in the 5-shot setting ((b) in Figure~\ref{fig:rq3_len}), its peak remains slightly below AR. By contrast, ``+dual cache'' reaches its steady performance early and maintains a nearly constant throughput (around 30 tokens/s) across sequence lengths, showing limited gains to generation length.

\subsubsection{Effect of batch size}

\begin{figure}[t]
    \centering
    \begin{subfigure}[b]{0.48\linewidth}
        \centering
        \includegraphics[width=\linewidth]{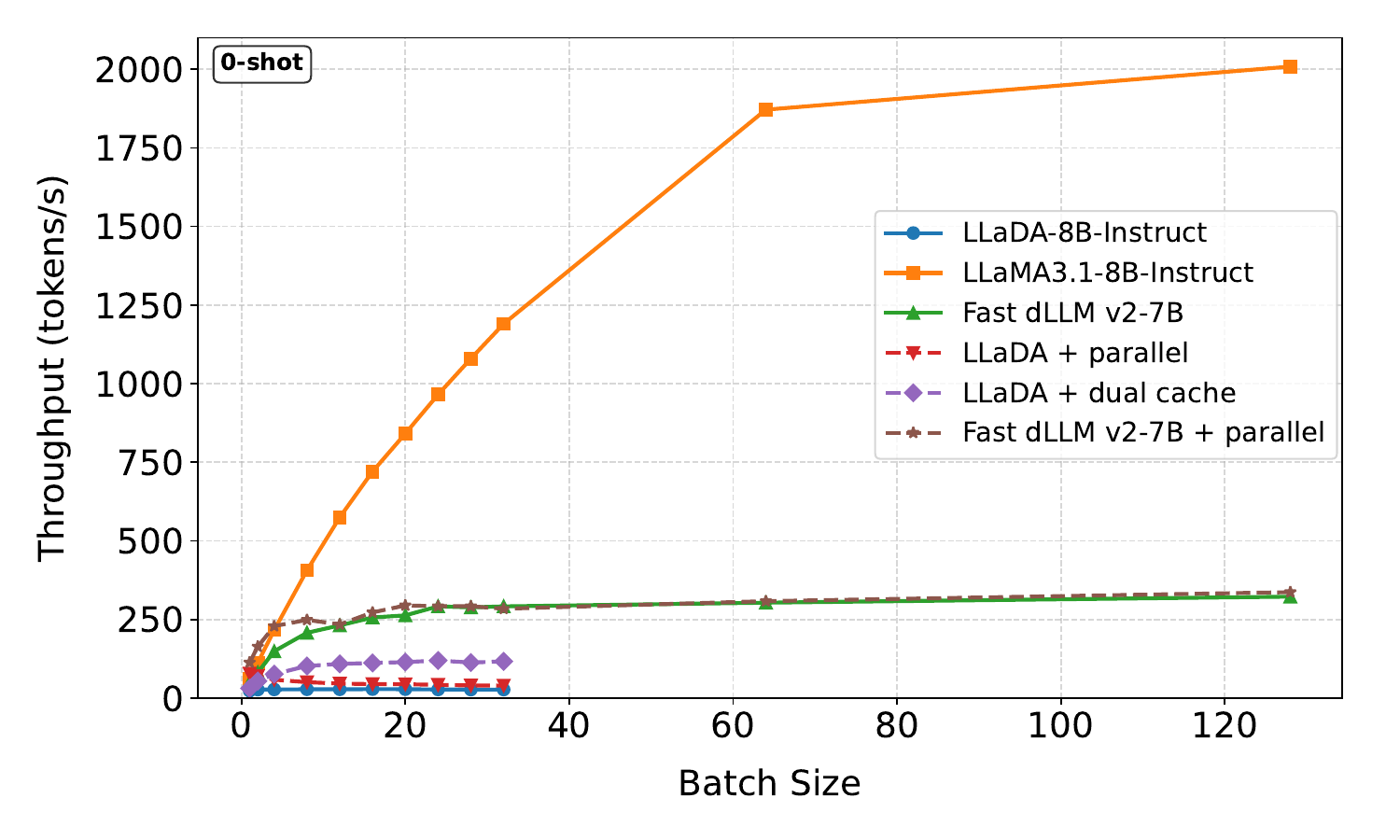}
        % \caption{Results under the 0-shot setting.}
        \caption{Throughput~(0-shot).}
        \label{fig:scaling_a}
    \end{subfigure}
    \hfill
    \begin{subfigure}[b]{0.48\linewidth}
        \centering
        \includegraphics[width=\linewidth]{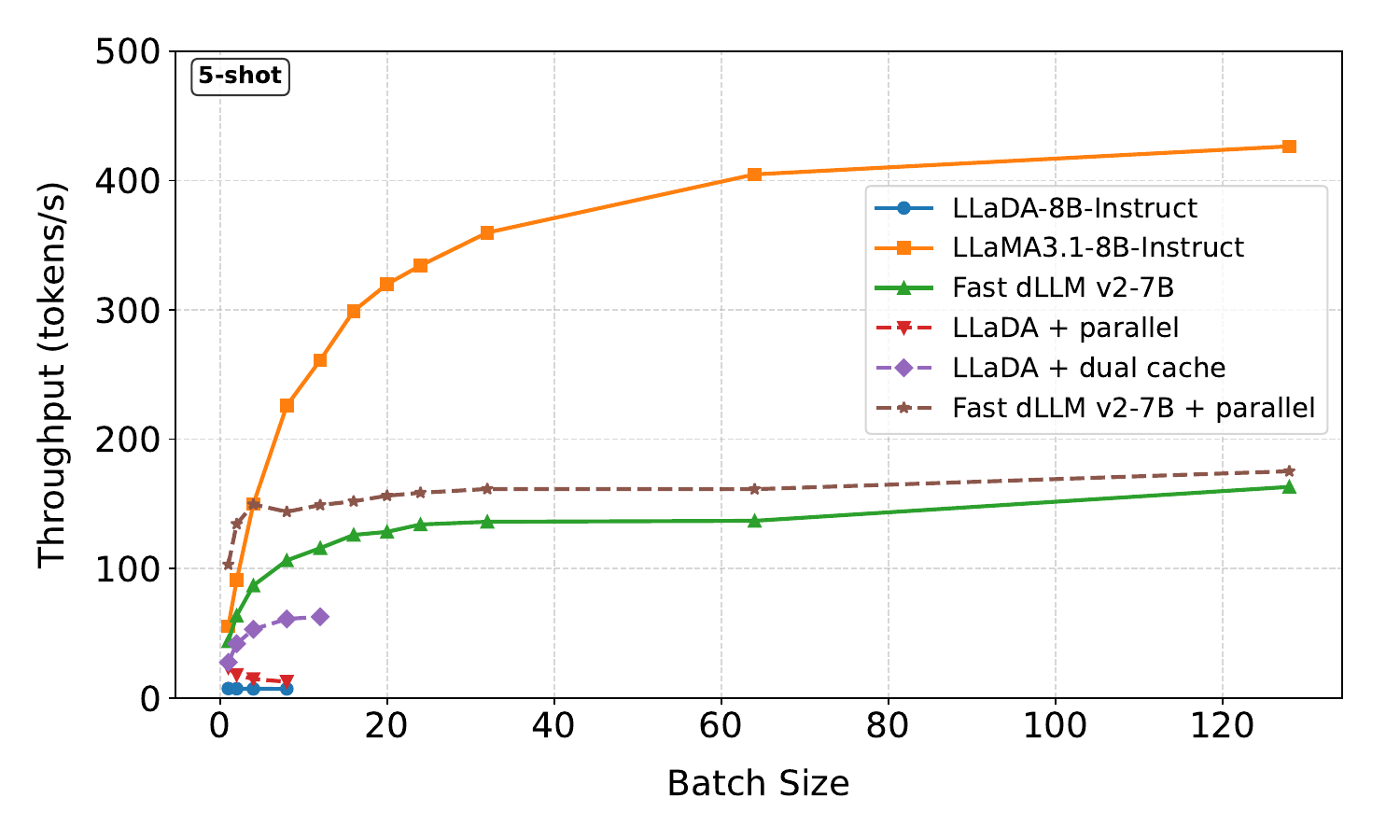}
        % \caption{Results under the 5-shot setting.}
        \caption{Throughput~(5-shot).}
        \label{fig:scaling_b}
    \end{subfigure}
    % \caption{(a) and (b) show the throughput comparison across different batch sizes under the 0-shot and 5-shot settings, respectively.}
    \caption{The throughput comparison across different batch sizes under the 0-shot~(a) and 5-shot~(b) settings, respectively.}
    \label{fig:rq3_bs}
\end{figure}
\vspace{-0.3cm}

\paratitle{Experimental Settings.} The experimental setup in this section follows the same batch size configuration as in Section~\ref{sec:rq1_effect}. Here, our goal is to examine how different acceleration strategies behave under varying batch sizes and to analyze their impact on overall inference efficiency. 

\paratitle{Main Results.} The results are presented in Figure~\ref{fig:rq3_bs}.

First, across all batch sizes, the benefit of block diffusion with the parallel strategy gradually diminishes as the batch size increases, at which point the AR model becomes more efficient. We observe that for block diffusion LLM with the ``+parallel'' strategy, throughput continues to improve as the batch size grows, but the rate of improvement is smaller than that of AR, and a clear turning point emerges. Specifically, when the batch size is small (\eg batch size of 2 for 5-shot and batch size of 4 for 0-shot), Block Diffusion LLM performs better than AR; beyond this point, the trend reverses. This behavior arises because increasing the batch size causes diffusion-based models to reach the compute-bound regime more quickly, thereby diminishing their advantage in parallel decoding. Finally, the ``+parallel'' strategy tends to converge to the same throughput as the original Block Diffusion LLM.

Second, the two acceleration strategies for LLaDA display distinct scaling behaviors as the batch size increases. Specifically, the ``+dual cache'' variant consistently improves throughput with larger batch sizes and generally outperforms the ``+parallel'' variant. In contrast, the throughput of ``+parallel'' steadily decreases as the batch size grows. In the 5-shot setting, ``+dual cache'' achieves higher throughput than ``+parallel'' across all batch sizes, while in the 0-shot setting, its advantage emerges only when the batch size exceeds 2. This trend suggests that the ``+dual cache'' mechanism scales more effectively under increasing batch-level workloads, though it ultimately hits memory limitations that lead to out-of-memory (OOM) failures at very large batch sizes.

% summary of rq3
% \begin{promptbox}[Summary]{blue!60!black}
\begin{tcolorbox}[colback=blue!5!white,colframe=blue!55!black,width=1.0\textwidth,title={Summary}]{
Overall, we find that these acceleration strategies are most effective when the batch size is set to one, but show marginal benefits as the batch size grows. Our main findings are summarized as follows:

\vspace{0.15cm}
\textit{$\bullet$ For block diffusion, when batch size is set to 1, the ``+parallel'' strategy yields the most substantial gains, enabling the fastest decoding speed and even outperform the AR baseline. As the batch size increases, however, AR becomes more efficient. }

\vspace{0.15cm}
\textit{$\bullet$ For vanilla DLMs (\eg LLaDA), acceleration performance depends strongly on prompt length at a batch size of one: with short prompts, ``+parallel'' achieves higher efficiency than ``+dual cache'' and even exceeds AR, whereas with long prompts, both strategies offer limited gains and remain slower than AR.}
}
\end{tcolorbox}

%% file: sections/future_directions.tex
\section{Future Directions}
\label{sec:future_directions}
In this section, we outline key future directions for DLMs, covering infrastructure, efficiency, adaptability, decoding, integration, and applications.

\paratitle{Comprehensive Benchmark for DLM Efficiency.}
As discussed in Section ~\ref{sec:evaluation_issues}, the current DLM community lacks standardized evaluation benchmark. It is important to establish consistent and comprehensive benchmarks for assessing DLM efficiency—including fair infrastructure setups, comprehensive experimental settings, and standardized evaluation metrics, to foster fairer comparison and drive the development of more efficient diffusion-based language models.

\paratitle{Inference Infrastructure for DLMs.}
Despite the impressive progress of DLMs, major machine learning ecosystems provide only limited optimization and deployment support for DLMs, posing practical challenges for researchers and developers. In particular, DLMs lack mature, open-source inference infrastructure akin to vLLM~\cite{18-sosp-2023-woosuk}, making efficient serving of DLMs difficult.

\paratitle{Optimization for Computational Efficiency.}
The efficiency of DLMs is affected by multiple factors, among which computational efficiency remains a major bottleneck. As discussed in Section ~\ref{sec:evaluation_analysis}, reducing the number of floating-point operations required to generate each token (FLOPs/token) is critical to improving the computational efficiency of DLMs. This can be achieved through two complementary approaches: The first approach focuses on reducing the computational cost of a single forward pass. Future work may explore more advanced caching techniques, as well as architectural optimizations such as sparse or linear attention, to further improve computational efficiency during each forward pass. In addition, the total diffusion sampling steps can be reduced to further lower the average FLOPs/token. Future work may explore better distillation-based approaches to decrease the number of iterations while preserving generation quality. Ultimately, combining these two optimization strategies could lead to multiplicative improvements in inference efficiency.

\paratitle{Variable-Length Generation in DLMs.}
Existing DLMs often adopt a predefined generation length, which often leads to redundant computation or suboptimal generation quality. Specifically, this static length allocation creates a critical trade-off: if the predefined length is too short, the model fails to complete complex tasks; if it is too long, excessive computation is wasted, often leading to degraded generation quality due to overextended denoising. Therefore, a more principled approach is to enable DLMs to adaptively control their generation length according to the difficulty of each sample, which represents an important direction for future research.

\paratitle{Decoding and Sampling Strategies in DLMs.}
An important future direction lies in improving decoding and sampling strategies to better align DLM inference with its pre-training dynamics. Current DLMs often rely on manually designed unmask/remask schedules and fixed denoising orders, which may lead to a mismatch between training and inference behaviors. Future research could explore adaptive masking policies that dynamically adjust the denoising trajectory based on token confidence, entropy, or contextual coherence. Such improvements would narrow the pretrain–inference gap and contribute to more stable, efficient, and controllable diffusion-based text generation.

\paratitle{Integration between DLMs and AR Models.}
Future work may explore tighter integration between diffusion and AR paradigms from both architectural and decoding perspectives. At the architectural level, future work could explore hybrid designs such as block diffusion, aiming to combine AR’s sequential precision with DLM’s parallel generation. At the decoding level, guidance-based inference, where AR models serve as structural or probabilistic guidance during DLM denoising, may further improve controllability and fluency.

\paratitle{Applications of DLMs.}
Although DLMs are generally less efficient than AR models, their parallel architecture and iterative refinement can offer advantages in compute-rich or latency-sensitive scenarios. We highlight two representative application directions: (1) on-device inference. Combining with parallel decoding techniques, DLMs can achieve higher arithmetic intensity and throughput efficiency, making them well-suited for latency-sensitive and low-concurrency environments such as in-vehicle systems, mobile devices, and edge computing platforms. (2) structured and non-sequential generation. DLMs exhibit notable advantages in structured or non-sequential generation tasks such as code synthesis, table generation, and logical planning~\cite{19-arxiv-2025-samar, 20-arxiv-2025-chengze}. This suggests that the DLMs hold strong potential for scenarios requiring global semantic planning or multi-constraint generation.

%% file: sections/conclusion.tex
\section{Conclusion}
\label{sec:conclusion}

In this report, we present a preliminary investigation into the efficiency of DLMs through both theoretical and empirical analyses. Our analysis reveals that existing evaluations remain limited in scope and consistency, lacking a comprehensive understanding of DLM efficiency across diverse conditions. In response, we introduce a theoretical analysis based 
% {on the roofline model}
{on the model-side and hardware-size computational efficiency}, revealing that while DLMs exhibit inherently high computational parallelism, they are often compute-bound with larger FLOPs per token, which constrains their practical efficiency. Empirical results further shows that acceleration strategies such as dual cache and parallel decoding bring substantial speedups under small-batch settings. Overall, these insights are expected to encourage the community to deepen the exploration of DLM architectures and acceleration strategies, fostering a more diverse and efficient ecosystem.

%% file: sections/appendix.tex
\newpage
\appendix

\section{Arithmetic Intensity Estimation}
\label{app:arint_est}
We estimate the specific expressions of arithmetic intensity for different model architectures, including AR, DLM, and block diffusion. The notations we will be using are as follows:\\
\begin{table}[h]
\centering
\renewcommand{\arraystretch}{1.4} % 控制行距
\setlength{\tabcolsep}{6pt}       % 控制列间距
\small
\begin{tabular}{cl}
\toprule
\textbf{Notation} & \textbf{Description} \\
\midrule
$N$       & Total number of parameters. \\
$B$       & Batch size. \\
$n_l$     & Number of layers. \\
$n_h$     & Number of attention heads. \\
$n_d$     & Dimension of each attention head. \\
$\alpha$  & Expansion ratio of the FFN layer. \\
$L$       & Sequence length. \\
$d$       & Hidden size. \\
$G$       & Block size. \\
\bottomrule
\end{tabular}
\label{tab:notations}
\end{table}

The mathematical estimation expressions for the arithmetic intensity of AR, Block Diffusion, and DLM are presented as follows:
\begin{equation}
\text{ArInt}_{AR}= \frac{BN}{N+Bn_ln_hn_dL}.
\end{equation}

\begin{equation}
\text{ArInt}_{Block Diffusion}= \frac{2Bn_l(2Gd^2+\alpha^2Gd^2+LGd)}{N+2Bn_ldL+Bn_ldG}.
\end{equation}

\begin{equation}
\text{ArInt}_{DLM}= \frac{2Bn_l(2Ld^2+\alpha^2Ld^2+L^2d)}{N+Bn_ldL}.
\end{equation}